\documentclass[review, 3p]{elsarticle}

\usepackage{lineno}
\usepackage[T1]{fontenc}
\usepackage{booktabs}
\usepackage{threeparttable}
\usepackage{multirow}
\usepackage{adjustbox}
\usepackage{rotating}
\usepackage[table]{xcolor}
\usepackage{float}
\usepackage[caption = false]{subfig}
\usepackage{graphicx}
\usepackage[pagebackref=true,breaklinks=true,colorlinks,bookmarks=false]{hyperref}
\usepackage{comment}
\modulolinenumbers[5]
\newcommand{\secref}[1]{\S\ref{#1}}
\journal{Image and Vision Computing}

\begin{document}

\begin{frontmatter}

\title{A Survey on Computer Vision based Human Analysis in the COVID-19 Era}

\author[iremaddress]{Fevziye Irem Eyiokur\corref{mycorrespondingauthor}}
\cortext[mycorrespondingauthor]{Corresponding author}
\ead{fevziye.yaman@kit.edu}
\author[hazimddress]{Alperen Kantarc{\i}}
\author[hazimddress]{Mustafa Ekrem Erak{\i}n}
\author[naserddress1,naserddress2]{Naser Damer}
\author[ferdaaddress]{Ferda Ofli}
\author[ferdaaddress]{Muhammad Imran}
\author[janezaddress]{Janez Križaj}
\author[aliaddress1,aliaddress2]{Albert Ali Salah}
\author[iremaddress,alexaddress]{Alexander Waibel}
\author[janezaddress]{Vitomir Štruc}
\author[hazimddress]{\\Haz{\i}m Kemal Ekenel}
\address[iremaddress]{Institute for Anthropomatics and Robotics, Karlsruhe Institute of Technology, Karlsruhe, Germany}
\address[hazimddress]{Department of Computer Engineering, Istanbul Technical University, Istanbul, Turkey}
\address[naserddress1]{Fraunhofer Institute for Computer Graphics Research IGD, Darmstadt, Germany}
\address[naserddress2]{Department of Computer Science, TU Darmstadt, Darmstadt, Germany}
\address[ferdaaddress]{Qatar Computing Research Institute, HBKU, Doha, Qatar}
\address[janezaddress]{Faculty of Electrical Engineering, University of Ljubljana, Tržaška cesta 25, 1000 Ljubljana, Slovenia}
\address[aliaddress1]{Department of Information and Computing Sciences, Utrecht University, Utrecht, the Netherlands}
\address[aliaddress2]{Department of Computer Engineering, Boğaziçi University, Istanbul, Turkey}
\address[alexaddress]{Carnegie Mellon University, Pittsburgh, United States}

\begin{abstract}
The emergence of COVID-19 has had a global and profound impact, not only on society as a whole, but also on the lives of individuals. Various prevention measures were introduced around the world to limit the transmission of the disease, including 
face masks, mandates for social distancing and  regular disinfection in public spaces, and the use of screening applications. These developments also triggered the need for novel and improved computer vision techniques capable of $(i)$ providing support to the prevention measures through an automated analysis of visual data, on the one hand, and $(ii)$ facilitating normal operation of existing vision-based services, such as biometric authentication schemes, on the other. Especially important here, are computer vision techniques that focus on the analysis of people and faces in visual data and have been affected the most by the partial occlusions introduced by the mandates for facial masks. Such computer vision based human analysis techniques include face and face-mask detection approaches, face recognition techniques, crowd counting solutions, age and expression estimation procedures, models for detecting face-hand interactions and many others, and have seen considerable attention over recent years. The goal of this survey is to provide an introduction to the problems induced by COVID-19 into such research and to present a comprehensive review of the work done in the computer vision based human analysis field. Particular attention is paid to the impact of facial masks on the performance of various methods and recent solutions to mitigate this problem. Additionally, a detailed review of existing datasets useful for the development and evaluation of methods for COVID-19 related applications is also provided. Finally, to help advance the field further, a discussion on the main open challenges and future research direction is given at the end of the survey. This work is intended to have a broad appeal and be useful not only for computer vision researchers but also the general public. 

\end{abstract}

\begin{keyword}
\texttt{Computer vision\sep COVID-19 \sep human analysis \sep masked faces \sep survey}
\end{keyword}

\end{frontmatter}


\section{Introduction\label{Sec: Introduction}}
The COVID-19 pandemic took the world by storm. Since the first large-scale outbreak in December $2019$ in Wuhan, China, COVID-19, a highly infectious atypical (viral) pneumonia caused by the zoonotic coronavirus SARS-CoV-$2$, spread throughout the globe and resulted in around $600$ million recorded cases and over $6.5$ million deaths by mid $2022$ according to information from Worldometer\footnote{Accessible from: \url{https://www.worldometers.info/coronavirus/}} \cite{batagelj2021correctly}. To contain the spread of the disease, minimize cases and limit the number of deaths, governments across the world started introducing prevention measures that had a profound impact on peoples' lives and changed their behavior and daily routines. Common prevention measures included (mandatory) face masks in public spaces, medical facilities and crowded areas, requests for social distancing, and restrictions on the allowed crowd size at different events, among others~\cite{mask_usage,mask_usage_2}. 

To help combat COVID-19, the computer vision community quickly took an active stance and initiated a wide range of research activities that resulted in novel techniques for COVID-19 detection and severity analysis from medical images~\cite{ulhaq2020covid,bhargava2021novel}, monitoring solutions for assessing compliance with the given prevention measures \cite{batagelj2021correctly, nguyen2021effectiveness,petrovic2022smart}, screening approaches for flagging potentially sick subjects \cite{hussain2021iot,tan2020application,tan2020fighting}, infection-risk assessment methods \cite{rezaei2020deepsocial}, and efficient biometrics-based authentication schemes tailored towards the characteristics of the COVID-19 era \cite{montero2021boosting,DBLP:conf/biosig/NetoBPSDS021}. These solutions have been swiftly adopted in practice and were observed to have a critical role in the  efforts towards containing the pandemic~\cite{AIforCOVID19}. They allowed to automate many monitoring tasks, improved situation-awareness and facilitated large-scale screening efforts. Furthermore, themed workshops, such as the International Workshop on Face and Gesture Analysis for COVID-19 (FG4COVID19)\footnote{URL: \url{https://fg4covid19.github.io/}}, were organized in the scope of major computer vision conferences to provide a platform for discussion and presentation of the latest vision techniques related to COVID-19.   
 
 A key component of many of the solutions discussed above are, what we refer to in this survey as, \textit{computer vision based human analysis} (CVHA) techniques that focus on the analysis of people and faces in visual data. While considerable progress has been made in the general area of vision based human analysis, the COVID-19 pandemic introduced several new challenges that have been underexplored in the literature before, e.g.:
 \begin{itemize}
     \item \textbf{Mask-based occlusions:} One of the globally most prevalent prevention measures, introduced in response to COVID-19, are face masks. The presence of face masks represents a considerable obstacle with an adverse impact on the performance of many CVHA techniques, such as facial landmarking, face detection and recognition, but also related (auxiliary) tasks such as face image quality assessment (FIQA), presentation attack detection (PAD) and others. Dedicated mechanisms are, therefore, needed to handle this type of occlusion. It is important to note that partial occlusions of the facial area have been studied also in the pre-COVID-19 era \cite{martinez2002recognizing,vstruc2010confidence,ekenel2009occl}. However, most of the research from that period was not focused specifically on face masks and, as a result, techniques developed for more general occlusions were observed to lead to suboptimal performance for many COVID-19 related human-centered vision tasks.        
     \item \textbf{Relevant datasets:} The majority of modern vision techniques relies on machine learning and is, hence, trained using suitably annotated training data. 
     Before the start of the pandemic, there was an obvious lack of  datasets suitable for the development of CVHA techniques for combating COVID-19. Especially datasets with masked faces (and people) with high-quality annotations were not widely available. Several factors contributed to the lack of such datasets: $(i)$ there was limited interest in vision problems involving masked-faces (and masked-people) only, $(ii)$ the occlusions caused by the face masks made it difficult to generate annotations of reasonable quality (e.g., facial landmarks, accurate bounding boxes, segmentation/parsing maps, etc.), and $(iii)$ a wide variety of facial masks with highly diverse appearance was introduced during the pandemic, but was not available in the pre-COVID-19 era.       
     \item \textbf{Ethical considerations and social impact:} While CVHA techniques for COVID-19 were developed with the goal of more efficiently combating the pandemic, the deployment of such techniques also raises ethical considerations and comes with a considerable societal impact. For example, installments of screening and monitoring applications can help to contain the spread of the coronavirus, but may also be extended into citizen surveillance and impact the privacy of individuals. 
 \end{itemize}
 
A considerable amount of work has been conducted over the course of the last three years to address the above challenges and has been covered partially in recent survey papers.  Wang \textit{et al.} \cite{Wang21}, for example, reviewed techniques for masked facial detection and associated datasets. Alzubi \textit{et al.} \cite{alzu2021masked} as well as Utomo and Kusuma \cite{utomo2021masked} discussed dedicated face recognition techniques for masked faces.
Elbishlawi \textit{et al.} \cite{elbishlawi2020deep} reviewed crowd-counting techniques and pointed to the importance of this technology for COVID-19.  Related to these works is also the survey of Ulhaq \textit{et al.} \cite{ulhaq2020covid}, which covers computer vision techniques applicable mostly to medical data, diagnostics and clinical management, and the deep-learning oriented review paper by Shorten \textit{et al.} \cite{shorten2021deep}, where vision approaches, again mostly related to medical applications, are briefly discussed. Although the listed works provide some insight into CVHA research related to COVID-19, they focus on specific problems only, e.g., masked face detection or recognition, or provide a partial picture of the broader (and interconnected) research area.  A well-structured and thorough survey on COVID-19 focused vision-based human analysis techniques, on the other hand, is, to the best of our knowledge, still missing from the literature.

\begin{figure}[t!]
     \centering
\includegraphics[width=0.99\linewidth]{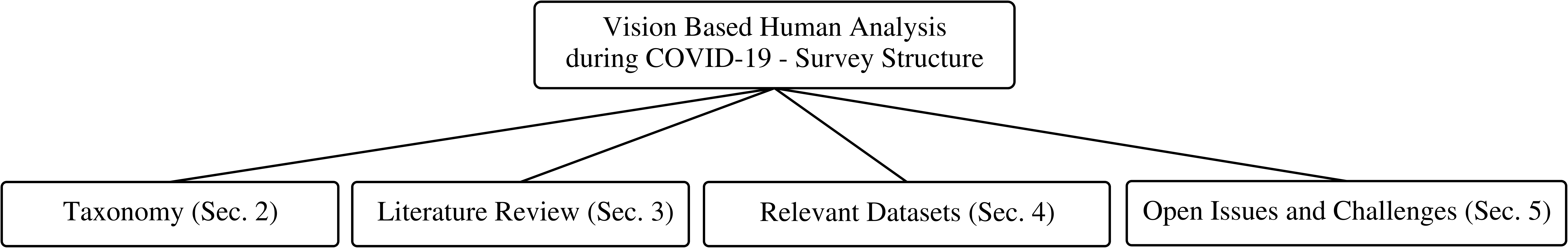}
\caption{High-level structure of the survey. We provide a comprehensive review of recent human-centered computer vision techniques for combating COVID-19, discuss existing datasets and data-generation procedures, and present a list and discussion of the most important open issues and future research directions.}
\label{fig:structure}
\end{figure}

In this work, we aim to address this gap and present a comprehensive overview of computer vision techniques that analyze visual data of people and faces with COVID-19 applications in mind. The goal of the survey is to: $(i)$ provide a high-level taxonomy and background on vision techniques applied to human analysis relevant to COVID-19~(Sec.~\secref{Sec: Background}) $(ii)$ present a consolidated summary of recent research activities in this area~(Sec.~\secref{Sec: Survey}), $(iii)$ provide a review of dataset collection and generation efforts~(Sec.~\secref{Sec: Datasets}), and $(iv)$ elaborate on open problems and challenges with the goal of providing a basis for future research activities~(Sec.~\secref{Sec5}). The overall structure of the survey is illustrated in Figure~\ref{fig:structure}. The work is primarily intended for researchers looking for a broad overview of computer vision research for COVID-19, but also other stakeholders interested in this topic.     

We make the following main contributions in this survey:
\begin{itemize}
    \item We present a comprehensive review of computer vision techniques that analyze imagery of people and faces to support the COVID-19 containment efforts and discuss over $200$ relevant references that cover diverse but relevant topics from this problem domain. 
    \item We provide a taxonomy of existing solutions for the most relevant tasks studied as part of vision-based research for COVID-19, e.g., face mask detection, masked face detection, masked face recognition, crowd analysis, etc.
    \item We discuss issues beyond the technological solutions, such as ethics, social impact and elaborate on open problems and future research directions.
\end{itemize}

Since COVID-19 will not be the last pandemic the world faces, we believe this survey will help technological preparedness for similar situations, and ultimately improve the robustness and usability of relevant technologies.

\section{Taxonomy on Computer Vision based Human Analysis (CVHA) for COVID-19 \label{Sec: Background}}
\begin{figure}[t!]
     \centering
\includegraphics[width=0.7\linewidth]{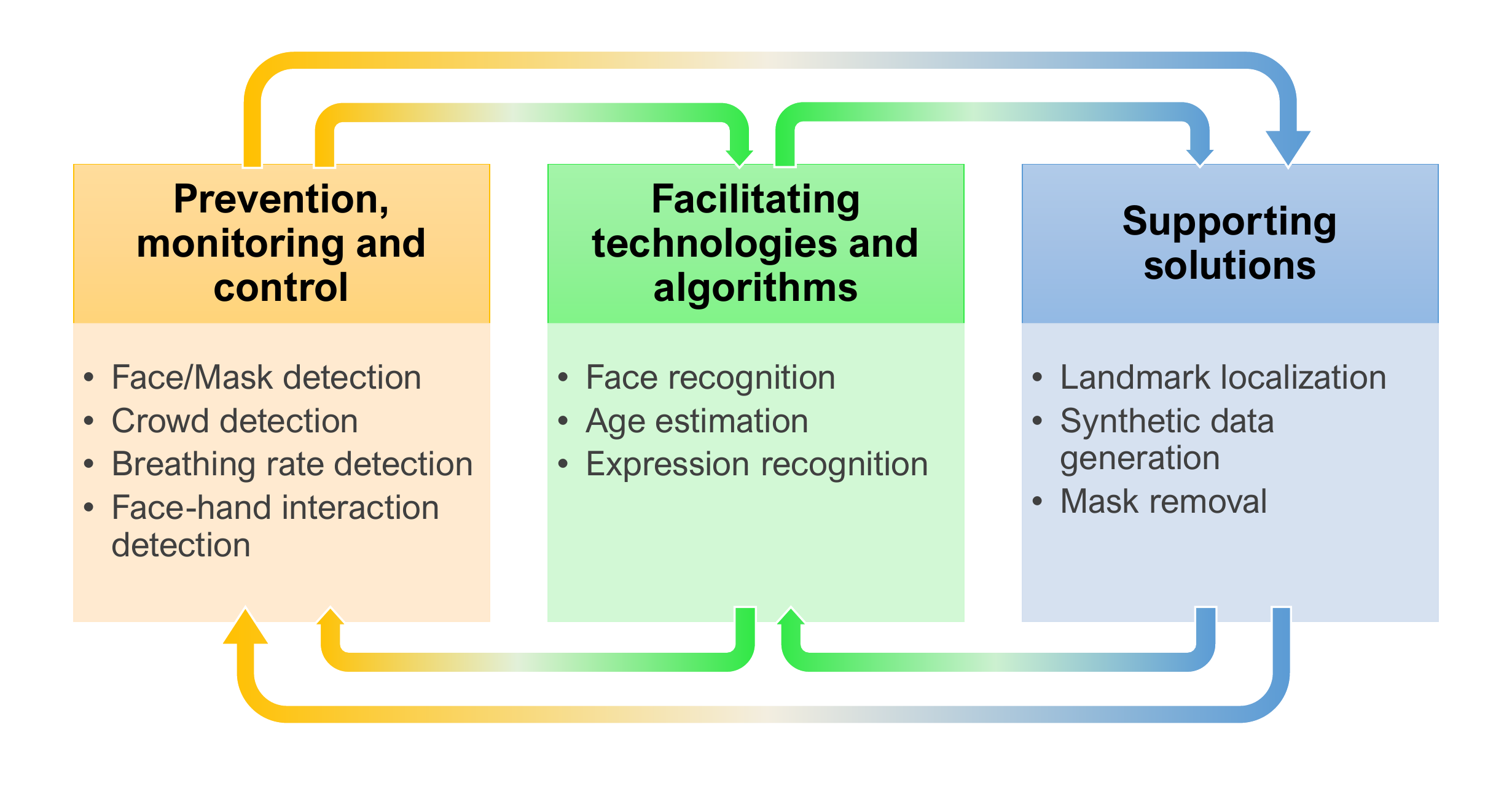}
\caption{High-level taxonomy of Computer Vision based Human Analysis (CVHA) techniques surveyed in this paper with respect to the targeted goal.}
\label{fig:taxCVHAtechniques}
\end{figure}

The COVID-19 pandemic triggered a need for efficient computer vision techniques related to different problems in visual human analysis that can broadly be categorized into three groups based on their overall goals, as also illustrated in Figure~\ref{fig:taxCVHAtechniques}, i.e.:
\begin{itemize}
    \item \textbf{Techniques for prevention, monitoring and control:} The goal of the first group of CVHA techniques is to help prevent the spread of COVID-19 and monitor compliance with the given prevention measures and typically aim to detect/analyze some characteristics (e.g., the presence of masks, the crowd size, or physiological changes/abnormalities) of the subjects in the visual data. Techniques from this group are applicable for screening purposes and as a source of critical statistical data for governments, health organizations and regulatory bodies. CVHA solutions covered in this survey from this group include face/mask detection algorithms~\cite{tomas2021incorrect,razavi2022automatic,hu2021covertheface},  crowd-counting solutions designed for COVID-19 characteristics~\cite{razavi2022automatic,petrovic2020iot,sathyamoorthy2021covid,yang2021vision,rezaei2020deepsocial}, breathing rate detection techniques~\cite{queiroz2021thermal} and face-hand interaction detection approaches~\cite{eyiokur2022unconstrained,beyan2020analysis}.     
    \item \textbf{Facilitating algorithms:} The second group of techniques represents solutions that facilitate applications that are not immediately related to COVID-19 prevention, but whose performance is affected by the external circumstance caused by the pandemic, such as, the presence of face masks. A typical example of such an application is biometric identity inference from facial images, where face masks have been observed to have a considerable adverse effect on the overall recognition accuracy~\cite{damer2020effect}. Many techniques and algorithms have, therefore, been proposed in the last few years to enable such critical applications also during COVID-19, but with minimal performance loss.  CVHA techniques from the  group of facilitating algorithms reviewed in this paper include face recognition solutions for masked faces \cite{anwar2020masked,damer2020effect,iccvrecognitionchallenge,wang2021mask,zhang2021arface,wang2021maskout}, as well age estimation \cite{yolcu2021multi,golwalkar2020age} and facial expression recognition approaches \cite{yang2020facial,abate2022limitations} that were all extended recently with the goal of improving robustness with masked faces.   
    
    \item \textbf{Supporting solutions:} The last group  of techniques in our taxonomy represents supporting solutions that do not address specific problems with real-world COVID-19-related applications, but are needed to enable techniques from the two groups above. The most important solutions from this group discussed in the survey are data generation techniques, capable of synthesizing artificial training data for the various computer vision models \cite{anwar2020masked,wang2021facex,wang2020masked}, mask removal techniques aiming to reconstruct the original (unocluded) facial images \cite{din2020novel,li2021face,coelho2021generative} and landmark localization (or/and alignment) techniques \cite{sha2021efficient,hu2021robust} that are used as preprocessing steps for other COVID-19-related CVHA solutions.    
\end{itemize}

We note that there is no clear separation between these three groups and there are clear interdependencies that are present in the presented taxonomy, as also highlighted in the overview Fig.~\ref{fig:taxCVHAtechniques}.

\section{Survey of CVHA techniques in the COVID-$19$ era \label{Sec: Survey}}

In this section, we summarize research on the different CVHA techniques that emerged during the COVID-19 pandemic and are covered in this survey. Specifically, we discuss research efforts focused on $(i)$ face/mask detection, $(ii)$ face recognition and various auxiliary tasks needed for face recognition systems, such as presentation attack detection and face quality estimation,  $(iii)$ facial expression recognition, $(iv)$ age classification, $(v)$ landmark localization, $(vi)$ crowd detection/counting, $(vii)$  breathing rate detection, $(viii)$ face-hand interaction detection, and $(ix)$ synthetic data generation.

\subsection{Face/Mask Detection}

Among the various prevention measures introduced around the world, face masks were likely the most wide spread and, in fact, were mandatory in various countries~\cite{mask_usage,mask_usage_2}. Facial masks  were also supported by the World Health Organization (WHO), who published a detailed guide on this topic~\cite{world2020advice}. Computer vision based detection techniques for masked faces are typically needed to monitor whether people comply with the advice of health organizations and governments in public spaces and to facilitate situational awareness. 

\begin{figure}[t!]
     \centering
\includegraphics[width=0.95\linewidth]{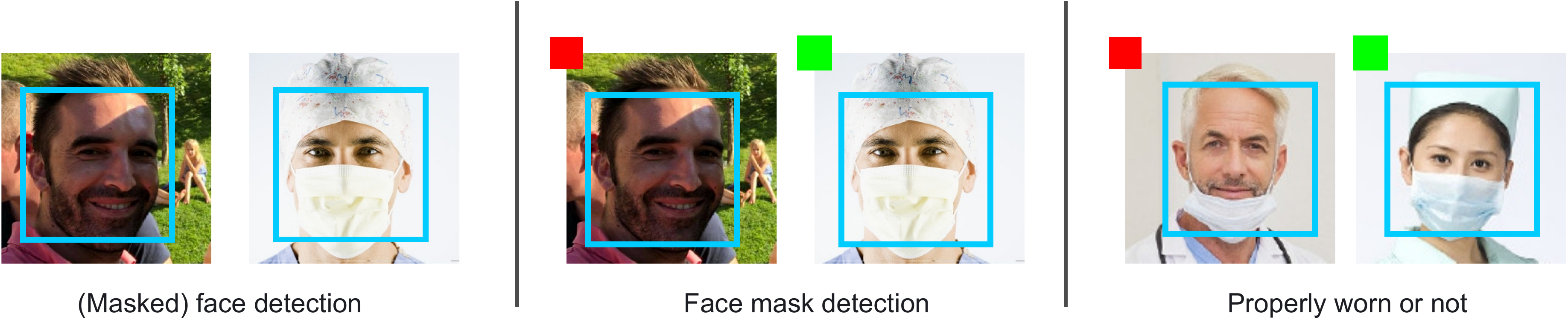}
\caption{Masked face detection is a specific object detection problem where the objects to be detected (i.e., faces) can appear both, with and without masks (left). The different extensions that appeared during the COVID-19 pandemic also incorporate decisions on whether the masks are present in the images (middle)  and whether the masks are worn correctly or not (right).}
\label{fig:face_detection_probs}
\end{figure}

In general, masked face detection is a specialized object detection problem where, in addition to the standard detection of non-occluded faces, the goal is to also reliably identify the presence of faces with masks in an image (or video frame). This task typically includes a binary decision (face present/face absent) for a given sub-region of the input image, which also defines the approximate spatial location of the (masked) facial area. Extensions of this problem that emerged during the pandemic, in addition to the (masked) face detection task, also often detect the presence of masks in the image (mask present/mask absent) or/and determine whether the mask is placed/worn in accordance with regulations and general guidelines or not~\cite{joshi2020deep, cabani2021maskedface, nagrath2021ssdmnv2, batagelj2021correctly, eyiokur2022unconstrained}, as shown in Figure~\ref{fig:face_detection_probs}.  

Before the pandemic, the masked face detection problem was mostly investigated as part of the more general detection tasks with partial occlusions, where the occlusions may have appeared due to the placement of the hands, the presence of sunglasses, gas masks, helmets, niqabs, and other objects that commonly cover some part of the face. In such settings, face detection methods are commonly observed to perform worse, with the performance degradations increasing as the occluded part of the face gets larger~\cite{yang2016wider}. One of the earliest pre-COVID-19 works on detecting masked faces was presented by Nieto-Rodrígue et al.~\cite{medicalmask_initial}, where the authors introduced a system that checks whether medical staff wears mandatory medical masks in the operating room. They used two distinct detectors: one for the face and the other for the medical mask. They employed the Viola-Jones object detector~\cite{viola2004robust} for both face and mask detection. They collected a dataset that contains faces with medical masks to train the detectors. Another work from this period~\cite{ge2017detecting} presented the first large-scale masked face detection dataset named MAFA and trained the locally linear embedding and convolutional neural networks (LLE-CNNs) for detecting faces with and without face masks. The proposed method first extracts (face) region proposals and describes them with a convolutional neural network (CNN). After this, a k-nearest neighbor (KNN) module refines the descriptors for recovering missing facial cues of masked faces. As the last step, a unified CNN is used to perform classification and regression to identify candidate facial regions and their overall positions in images/frames.

In~\cite{fan2021retinafacemask}, the authors introduced a refined version of the MAFA~\cite{ge2017detecting} dataset, called MAFA-FMD, that included only images with medical face masks. Using this new dataset, the authors proposed a novel context attention module to extract highly descriptive contextual features, such as face mask wearing conditions, and showed that the proposed approach outperforms the benchmark RetinaFace~\cite{deng2020retinaface} and YOLOv3~\cite{yolov3} face detectors. 
In~\cite{nagrath2021ssdmnv2}, Nagrath et al. introduced the SSDMNV2 system, which combines a single-shot multibox detector framework and a MobileNetV2~\cite{mobilenetv2model} based classifier for the detection of masked faces as well as face mask detection. This lightweight model is suitable for deploying on embedded devices and for real-time data processing. 
Another work that uses a single-stage face detector is~\cite{loey2021fighting}. Here, the authors used the YOLOv2 model~\cite{yolov2} to detect masked faces and ResNet-50~\cite{resnetmodel} to detect face masks. 
To overcome the challenge of scarce labeled data, Cabani et al.~\cite{cabani2021maskedface} built a synthetic dataset of masked faces to train robust face detection and face-mask detection models. The authors tried to imitate different mask-wearing conditions by using realistic image synthesis methods. However, they only used a single type of medical mask to simulate different wearing conditions, therefore, raising questions on the generalization capabilities of their models to real-world data, where facial masks may have different colors, shapes, and textures. 
Joshi et al.~\cite{joshi2020deep} proposed a framework to detect face masks from a video stream by using the MTCNN~\cite{mtcnn_paper} face detection model and classifying mask presence with MobileNetV2~\cite{mobilenetv2model}. They tested their framework on actual footage of public spaces, captured during the COVID-19 pandemic. The dataset contains multiple geographical locations and people from different ethnicities and the proposed method was demonstrated to outperform RetinaFaceMask~\cite{fan2021retinafacemask} on the considered dataset. The authors of~\cite{wang2021hybrid} proposed a two-stage Faster R-CNN~\cite{ren2015fasterrcnn} network with an InceptionV2~\cite{inceptionv3model} model along with a novel wearing mask detection (WMD) dataset to address the masked face detection task. Through comprehensive experiments, they show that the two-stage detector provides a good trade-off between accuracy and computational complexity. 
The work of Roy et al.~\cite{roy2020moxa} proposed using a YOLOv3~\cite{yolov3} model along with the Single Shot MultiBox Detector (SSD)~\cite{liu2016ssd}  and Faster R-CNN~\cite{ren2015fasterrcnn} for masked face detection. The experimental results on the novel Moxa3K benchmark dataset~\cite{roy2020moxa} showed that YOLOv3~\cite{yolov3} achieves better  
performance than competing models while having comparable runtime.

In~\cite{eyiokur2022unconstrained}, Eyiokur et al. studied an extended detection problem, where each face image was classified into one of three classes: no mask, mask, and incorrectly worn mask. The authors introduced a labeled large-scale face mask detection dataset and using the newly collected data trained a RetinaFace model~\cite{deng2020retinaface} for face detection. Next, they employed several CNN models, namely, Inception-v3~\cite{inceptionv3model}, MobileNetV2~\cite{mobilenetv2model}, EfficientNet~\cite{efficientnetmodel}, etc., to classify the detected and cropped faces into the three above-mentioned classes. The authors also extensively tested their models, both on the proposed dataset as well as on other available datasets from the literature. Cross-dataset evaluations showed their dataset's representation power, which is crucial for new face mask detection datasets.
A similar problem was also studied in~\cite{jiang2021real}, where Jiang et al. proposed the Squeeze and Excitation (SE)- YOLOv3~\cite{yolov3} mask detector for the detection of properly worn masks. The main idea behind the approach was to integrate the SE block with the YOLOv3~\cite{yolov3} model to teach the network to focus on the crucial features. The authors also utilized a focal loss to solve the extreme foreground-background class imbalance. Experimental results showed that the proposed network achieved better localization and detection performances than competing models on the considered dataset. 
Kantarci et al.~\cite{biasaware2022kantarci} introduced a novel face mask detection dataset named Bias-Aware Face Mask Detection (BAFMD) dataset. The dataset has been collected using Twitter images with a specific focus on mitigating dataset bias for ethnicity, age, and gender. In order to reduce such biases, their dataset contains real-world face mask images with a more balanced distribution across different demographics, e.g., gender, race, and age. They train a YOLOv5~\cite{yolov5} object detector, which shows superior performance over other detectors.
In~\cite{batagelj2021correctly}, Batagelj et al. compare different masked face detectors and correct face-mask placement classification networks in detail using a dataset that they created using the MAFA~\cite{ge2017detecting} dataset. The reported experimental results provide insightful performance comparisons of various methods and show that the RetinaFace~\cite{deng2020retinaface} model is the most stable masked face detection model among the considered techniques. Furthermore, the authors demonstrated that all face detection models' performance deteriorate significantly, if face masks are present in the image as compared to faces without masks.
In~\cite{qin2020identifying}, the authors proposed a face mask-wearing identification method by combining image super-resolution and classification networks (SRCNet). They used a standard face detector for detecting and cropping faces with and without masks. After the detection step, the authors evaluated the image size to choose the next step. If an image's resolution was smaller than $150\times150$ pixels, i.e., the width or length was below 150 pixels, they applied the super-resolution model to enhance the high-frequency details of the image. If the image was already larger than $150\times150$ pixels, they skipped the super-resolution and subjected the image to a face mask-wearing classification network, which classify the mask-wearing conditions into one of three classes: no mask-wearing, incorrect mask-wearing, and correct mask-wearing. The reported experimental results show that applying super-resolution to low-resolution face crops boosts classification performance and that the presented model yielded competitive performance overall.

Most of the  methods proposed for (masked) face detection and related problems, such as facial mask detection, build on advances made in the generic object detection problem domain. However, to adapt/extend the existing detectors to work reliably with partially occluded face data or, specifically, with masked faces, these solutions incorporate minor modifications to the overall detection pipelines and, more importantly, introduce new, large-scale datasets with masked faces that contain diverse data with rich appearance variations induced by facial masks for model training. Due to the importance of these datasets for the masked face detection problems, they are discussed separately in Section~\ref{Sec: Datasets}.

\subsection{Face Recognition}

Similarly to face detection, where the appearance and widespread usage of facial masks had an adverse impact on the performance of existing face detection models, face recognition is another area, where facial masks negatively impacted the applicability of face recognition technology. 
In this section, we therefore provide an in-depth discussion of the effect of face masks on different components of face recognition systems and then review the efforts made so far to mitigate such negative effects. 

\subsubsection{The effect of wearing masks on face recognition}

Face recognition deployability is strongly affected by biometric sample capture and presentation, most prominently, face occlusions.
Face recognition in the presence of occlusions has been studied widely within the computer vision community~\cite{ekenel2009occl, DBLP:journals/pami/KimCYT05,OU20141559,DBLP:conf/iccv/SongGLLL19,9495272,DBLP:journals/access/NetoPBDSC22,OCFR2022, erakin2021occl}. However, most of the pre-COVID-19 work targeted general unstructured face occlusions.
The effect of the specific occlusion induced by face masks gained attention at the start of the COVID-19 pandemic.
An early work by Damer et al.~\cite{damer2020effect} evaluated the verification performance drop of face recognition systems when verifying unmasked-to-masked faces, in comparison to verifying unmasked faces, all with real masks and in a collaborative environment. This was followed by an extended study~\cite{damer2021extended} with a larger database and evaluation of both synthetic and real masks. 
As a part of the ongoing Face Recognition Vendor Test (FRVT), the National Institute of Standards and Technology (NIST) has released results (FRVT -Part 6A) on the effect of face masks on the performance of face recognition systems provided by vendors~\cite{ngan2020ongoing}. The results revealed that the verification accuracy with masked faces declined substantially. However, the study used simulated masked images under the assumption that their effect is representative of the effect of real face masks.
Following NIST's evaluation, the US Department of Homeland Security conducted a similar evaluation, however, on more realistic data~\cite{DHS-Masks-2020}. They also identified a significant negative effect of facial masks on the accuracy of automatic face recognition solutions.
A general conclusion by these studies was that the effect of masks was bigger on genuine pairs' decisions, in comparison to imposter pairs' decisions.
A study comparing the effect of face masks on the human experts/verifiers in comparison to automatic face recognition models concluded with a set of comments on different aspects of the correlation between the verification performance of humans and machines~\cite{humanFr}. The study showed a trend in the human experts' verification performance drop similar to that of automatic face verification models~\cite{humanFr}. In the next section, an overview of the solutions to mitigate this negative effect on face recognition performance is presented.

\begin{figure*}[t!]
    \centering
    \includegraphics[width=0.990\linewidth]{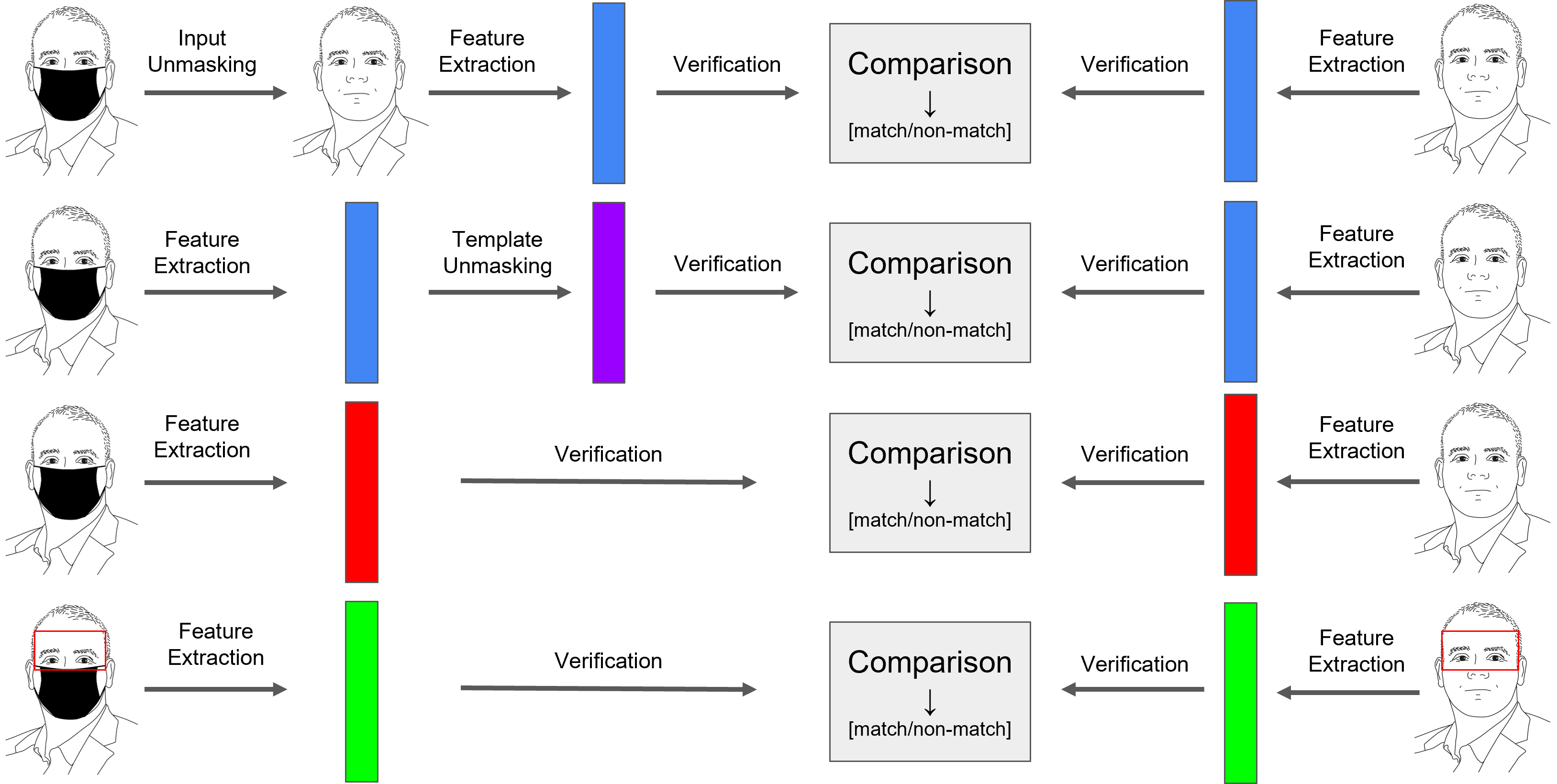}
    \caption{Masked face recognition solutions mainly fall under one of the four categories presented in this figure. From top to bottom, these solutions are face in-painting, template unmasking, model optimization, and periocular recognition. Blue rectangles are general-purpose face recognition models, purple rectangles are template unmasking models, red rectangles are face recognition models trained specifically to tolerate masked faces, and green rectangles are periocular recognition models.   }
    \label{fig:MFRS}
\end{figure*}

\subsubsection{Enhancing masked face recognition}

As validated by the studies discussed above, wearing a face mask does significantly affect the performance of face recognition technology. This by itself is intuitive, as the mask covers part of the facial information that face recognition models can use to discriminate between individuals. 
However, the insights from the discussed studies also inspired many innovative solutions aiming at enhancing the performance of masked face recognition. In this work, we present an operational categorization of these solutions based on their conceptual modeling of the masked face recognition problem. 
These solutions can be categorized into four groups, (a) mask in-painting, (b) template unmasking, (c) model optimization, and (d) periocular recognition,  and are presented in the following sections along with the main works that made significant contributions under each category. A graphical representation of these categories is also presented in Figure~\ref{fig:MFRS}, where masked face probes are processed in four different processes to be compared to an unmasked face reference.

\paragraph{(A) Mask in-painting} Under this category, illustrated in the top of Figure \ref{fig:MFRS}, the main goal is to detect and in-paint the face area covered by the mask before processing the face with conventional face recognition models. Such a process will not necessarily add additional identity-specific information to the face, because such in-painting processes are trained to predict the occluded area details from the visible parts of the face, and thus they extract the initial identity information from the already visible parts of the face. However, such in-painting will transfer the image into a distribution (domain) that is more similar to what general-purpose face recognition models are trained for and bring it closer to the unmasked reference. Seen as a domain adaption process, this can have the potential in enhancing the performance of masked face recognition. The main advantage with in-painting based strategies is the possibility to maintain the use of well-performing general-purpose face recognition models. The main disadvantage is that the training of the generative in-painting process is commonly expensive in terms of the required training data and the training computational cost~\cite{DBLP:conf/cvpr/LiLY017}. Such generative processes might also result in artifacts that are out of the normal face image distribution compared to the original masked faces themselves~\cite{DBLP:conf/aaai/SongCSHH19}. 

Although face in-painting is in general a well-studied field with the recent methods producing photo realistic images~\cite{DBLP:journals/spl/HanW21,DBLP:journals/symmetry/NiuLLMY20,DBLP:journals/pr/ZhangWSYLKLZLYS22}, using this technique to enhance masked face recognition is still under-explored. Jiang et al.~\cite{jiang2022mask} recently addressed the specific issue of in-painting face mask areas without evaluating the effect on face recognition performance.
Such aesthetic-driven face in-painting of the mask area, i.e. mask removal, is discussed more in details in Section \ref{SubSec:simulated}. Similar in-painting approaches have been shown before to be beneficial, to some degree, in enhancing the recognition performance~\cite{Li2020} of occluded faces. 

\paragraph{(B) Template unmasking} Under this category, illustrated in the second row in Figure~\ref{fig:MFRS}, the main goal is to transfer the extracted masked face template into a form where it behaves similarly to a template extracted from an unmasked face of the same identity. Here, both the masked probe and the unmasked reference are processed by a general-purpose face recognition model, however, the masked face template typically undergoes another processing step.
Just like with in-painting, this process will not add identity information to the masked face template, but it rather will remove the template artifacts introduced by the mask information. The main advantage of such solutions is that it maintains the use of the well-performing general-purpose face recognition models and that the overhead computational cost of the template unmasking model is relatively negligible~\cite{boutros2022self} when compared to the face recognition model itself or the generative in-painting model discussed under the first category.

Despite the clear operational benefits of this category of solutions, relatively few works targeted such a concept. The first to do so was Boutros et al.~\cite{boutros2022self} that proposed to train a template unmasking model on top of any general-purpose face recognition model to transfer the masked face template to a form that behaves similarly to an unmasked face template of the same identity in the comparison operations. Based on the fact that the genuine comparisons are significantly more affected than the imposter ones when comparing masked to unmasked faces, the authors proposed the self-restrained triplet loss that assigns higher importance to positive pairs during training when the negatives pairs are deemed relatively distanced enough. Following a similar operational concept, a recent study also proposed to process the masked face template in a framework that utilizes contrastive learning~\cite{neto2021focusface}.

\paragraph{(C) Model optimization} Under this category, illustrated in the third row in Figure \ref{fig:MFRS}, the main goal is to train a face recognition model that can produce comparable embeddings for both masked and unmasked faces. Understandably, training such a solution requires having masked and unmasked face samples in the training data. This  also requires, in part, general-purpose face recognition training goals such as direct embedding learning~\cite{DBLP:conf/cvpr/SchroffKP15} or embedding learning through classification~\cite{DBLP:journals/corr/abs-2109-09416,deng2019arcface}. The main advantage of this category of solutions is that both the masked and unmasked faces are processed with the same model. However, such solutions induce the need for a tedious training process and considerable amounts of training data that  also needs to include masked faces. Additionally, including masked faces in the training process might render the resulting model less accurate when comparing pairs of unmasked faces~\cite{deng2021mfcosface}. 
This shortfall was recently targeted in the literature with a high degree of success~\cite{huber2021mask}, where the authors forced the face recognition model to produce optimal templates for both, masked and unmasked faces by incorporating a template-level knowledge distillation loss between the trained network and a general-purpose face recognition network.

Most of the works addressing masked face recognition so far fall under this category. The authors in~\cite{montero2021boosting} combined the ArcFace loss~\cite{deng2019arcface} with a mask-usage classification loss and noted it as  Multi-Task ArcFace~\cite{montero2021boosting}. Other work combined the traditional triplet loss and the mean squared error in an effort to improve the face recognition robustness to masks~\cite{DBLP:conf/biosig/NetoBPSDS021}. The authors in~\cite{deng2021mfcosface} theorized that the masked face recognition process requires a larger penalty margin when using the cosine loss. Others proposed improving the face template consistency using a pairwise loss~\cite{DBLP:conf/iccvw/QianZJCX21}. Geng et al.~\cite{geng2020masked} proposed to enhance masked face recognition performance through mask-like generative augmentation. Hsu et al.~\cite{9738595} experimented with different loss functions to determine their suitability for masked face recognition.
A hybrid backbone of residual block and self-attention components was proposed by~\cite{DBLP:conf/iccvw/ChangTL21}, an aspect that was also investigated in~\cite{DBLP:journals/apin/LiGLL21}. 

\paragraph{(D) Periocular recognition} Under this category, illustrated at the bottom of Figure \ref{fig:MFRS}, the main goal is to simply reduce the face recognition problem to be a partial face recognition problem. This assumes that the mask commonly covers the lower part of the face and maintains the visibility of the upper part of the face. This area that includes the eyes and the adjacent regions is commonly called the (peri)ocular region~\cite{DBLP:conf/icpr/NguyenRRD20}. The biometric literature refers to the recognition of this area, when the iris is not exclusively targeted, as periocular recognition~\cite{DBLP:journals/prl/Alonso-Fernandez16}. Periocular recognition can include the periocular region of one eye for some applications~\cite{DBLP:journals/sensors/BoutrosDRKK22,DBLP:journals/ivc/BoutrosDRRKK20}. However, in the masked face recognition scenario, both right and left periocular regions are typically considered.

A number of works proposed to crop the masked face and focus the recognition task on the periocular region when the mask is present~\cite{9619841,9607897,9667012}. The need to use the periocular region for recognition purposes when faces are masked was extensively studied in~\cite{DBLP:journals/corr/abs-2203-15203}, including a detailed survey on periocular recognition technologies.

\subsubsection{Masked face recognition competitions}

Two major competitions were organized in an effort to attract novel solutions for masked face recognition. The first was the MFR2021 Masked Face Recognition Competition~\cite{MFRijcb} organized as part of the International IEEE Joint Conference on Biometrics (IJCB) 2021~\cite{ijcb2021}. The competition examined the deployability of the solutions by considering the compactness of the face recognition models. 
A private dataset was used for evaluation. The dataset contained real masked faces and represented a collaborative capture scenario. 
Out of 18 submitted solutions, 10 were able to outperform the widely used ResNet-100 baseline~\cite{resnetmodel} trained using the ArcFace loss~\cite{deng2019arcface}.
Most of the competition entries used synthetic or/and real masked face images in the training of their solutions.

The second competition was the Masked Face Recognition Challenge~\cite{iccvrecognitionchallenge} organized within the Face Bio-Metrics Under COVID? Masked Face Recognition (MFR) Workshop, one of the IEEE/CVF International Conference on Computer Vision Workshops~\cite{DBLP:conf/iccvw/2021}.
The competition included three test sets and used an online model testing system and provided a detailed evaluation of the submitted face recognition models.
The results of the competition pointed to the effectiveness of augmentation strategies simulating facial masks when training recognition models for the targeted task of masked face recognition.

\subsubsection{Masks and face recognition subsystems}

\paragraph{Presentation attack detection}

Presentation attacks on face recognition systems involve the presentation of an artifact or of human characteristics to a biometric capture subsystem in a fashion intended to interfere with system policy, as defined in ISO/IEC 30107-3~\cite{iso30107-3}. 
This can include attacks like face morphing~\cite{ICBMorphing,DBLP:journals/corr/abs-2203-06691}, makeup attacks~\cite{makeupattack}, or even identification circumvention attacks~\cite{circumvention}. However, given the attack scenarios, the attack that is most related to wearing masks is the spoofing attack, where an attacker presents an artifact to a biometric capture subsystem with the aim of impersonating a different identity~\cite{iso30107-3}.
Presentation attack detection solutions (PAD) aim at differentiating between non-attack samples, i.e. bona fide, and spoofing presentation attacks~\cite{PADsurv}. 
Such solutions can be based on user challenge (user performing a specific task/move), on special sensor characteristics, e.g. light field camera or thermal sensor, or on software solutions~\cite{PADsurv}.
The most widely spread software solutions depend on analyzing samples captured in the visible domain given the high deployability of visible-spectrum cameras in personal devices.
Such solutions can be texture-based~\cite{texturePAD}, motion-based~\cite{motionPAD}, frequency-based~\cite{freqPAD}, or a combination of two or more of these technologies~\cite{hybridPAD}.

Wearing a face mask changes the nature of the sample processed by the face PAD algorithm.
This was apparent in the wide-spread reporting of malfunctioning face logins into personal devices at the start of the COVID-19 pandemic, not only because of failing to match to the unmasked reference image, but also because the masked face is seen as a spoofing attack by the PAD algorithm.
This interesting fact was revealed by an extensive study presented by Fang et al.~\cite{FangPR_PAD} where the authors collected a set of unmasked and masked bona fide and attack samples and tested both the vulnerability of face recognition to such attacks and the performance of established PAD algorithms when processing masked attacks.
The study also presented a novel kind of attack where the attacks, printed or shown on a screen, were covered with a real mask.
The main findings of the study pointed out that PAD algorithms classify many of the masked bona fide samples as attacks.
The study also found that face recognition algorithms are still vulnerable to masked face attacks, especially when a real face mask is placed on the attacks~\cite{FangPR_PAD}.
An effort to reduce this effect on PAD performance was successfully presented in~\cite{fang2021partial} where the authors propose to train the PAD using partial pixel-wise labels, where the real masks placed on the attacks are considered to be a bona fide area in an attack sample. This was also supported by giving the non-covered parts of the face a higher influence in the PAD decision inference from the image, bringing the PAD behavior on masked faces closer to that of the unmasked faces~\cite{fang2021partial}.
Further efforts are required though to build publicly available masked face attack databases and mask-invariant masked face PAD algorithms.

\paragraph{Quality assessment} Face image quality (FIQ) measures the utility of an image to face recognition algorithms~\cite{ISOIEC29794-1,best2018learning}. 
This utility is measured with an FIQ score as defined in ISO/IEC 2382-37~\cite{ISOIEC2382-37}.
Various methods have been developed for face image quality assessment (FIQA) weather by building quality pseudo labels and learning to predict such labels~\cite{faceqnet,sddfiqa}, by measuring different aspects of face recognition model response to the investigated image~\cite{SERFIQA,MagFace}, or learning to predict the relative classifiability of a face by predicting its class-relative placement in a face recognition training process~\cite{CRFIQA}. 
As FIQA measures the utility to face recognition algorithms, it does not necessarily reflect the perceived image quality (IQ) measured by conventional general image quality assessment (IQA) solutions~\cite{WACV22_IQAvsFIQA,Quality_pose}.
However, IQ measures have been found to correlate to the face image utility, though to a much lower degree than FIQ~\cite{WACV22_IQAvsFIQA}.
As mentioned earlier, wearing a mask does lower the accuracy of face recognition, and thus it is expected also to be reflected in a lower FIQ.
This issue was investigated by Fu et al.~\cite{fu2021effect} where it was shown that even when the perceptual quality and capture environment do not change, the FIQ drops substantially when a mask is worn. 
This consistently correlates with the drop in face recognition performance, whether by machine or human experts~\cite{fu2021effect}.
Additionally, the networks performing FIQA did shift their attention away from the mask region and more towards the visible face region, more specifically the ocular region, as demonstrated in~\cite{fu2021effect}.

\subsection{Facial Expression Recognition}

Another important application affected by the presence of face masks is Facial Expression Recognition (FER).
FER is a longstanding computer vision problem, where the goal is to recognize specific facial expression or emotional states based on changes in facial appearance. It is generally acknowledged that different parts of the face are involved when expressing different expressions, as evidenced, for example, by the facial action unit coding (FACS) system, one of the most widely conceptual frameworks to the FER problem~\cite{cohn1999automated,valstar2006fully,zhi2020comprehensive,ekman1978facial}. Thus, occlusions of these areas, as caused, for example, by facial masks lead to obvious performance degradations. 

The problem of facial expression recognition under the presence of face masks was explored by Abate et al. in~\cite{abate2022limitations}. Here, the authors studied class activation maps (CAM) for different expressions and found that anger, happiness, sadness, and neutral expressions are most heavily represented around the nose and mouth areas. As results of this observation the authors concluded that FER models struggle to extract informative features from the face images when face masks are present. To address this problem two common mitigation strategies were proposed in the literature, i.e.: $(a)$ collecting/generating masked datasets with facial expressions that can be used for fine-tuning of existing FER models, and $(b)$ designing new models capable of performing facial expression recognition despite the presence of masks. 

\paragraph{(A) Datasets and fine-tuning} 

Collecting and labeling facial expressions is a difficult, time- and labor-intensive task that might also be subjective. The difficulty of labeling facial expressions carries over to the problem of Masked Facial Expression Recognition as well. Since there are no  datasets publicly available for this task, generally, simulated masks are utilized in the literature. Yang et al.~\cite{yang2020facial}, for example, developed a mask simulation method that uses facial landmarks and their orientations to fit a mask. They also annotated 13000 images from the Labeled Faces in the Wild (LFW) dataset~\cite{huang2008labeled} for facial expression recognition and compiled a new dataset, called LFW-FER. Finally, using the mask simulation methodology on the LFW-FER dataset, they generated a synthetic dataset for FER containing simulated mask, called M-LFW-FER that is publicly available for research purposes and can be used to fine tune FER models for expression recognition under the presence of facial masks. 

Similar ideas were also pursued by other works. Barros et al.~\cite{barros2021only}, for example, first detected the facial landmarks on images from the AffectNet dataset~\cite{mollahosseini2017affectnet} and fit a mask to the faces covering all the landmarks below the nose. Using the resulting MaskedAffectNet dataset, the authors then applied different training strategies, e.g., transfer learning, to their FaceChannel model~\cite{barros2020facechannel} to account for the presence of face masks.  When the authors trained the FER model from scratch using the MaskedAffectNet dataset, the model performance drastically deteriorated for unmasked applications. However, when the model first pretrained on a standard dataset and later fine-tuned, the FER performance was affected only so slightly, making it useful for both masked and unmasked facial images. 

\paragraph{(B) Mask-agnostic FER} Techniques from the second group aim to design models that are robust (agnostic) with respect to the presence of facial masks and perform similarly for masked and unmasked faces. Yang et al.~\cite{yang2021face} developed a new approach for masked face expression recognition along these lines. Their model consists of two parts. The first part includes a classifier for masked and unmasked recognition that generates a binary attention heatmap for the face masks. The second part of the model takes the binary attention heatmaps and convolutional face features to classify the facial expression.
The authors show that their model outperforms other state-of-the-art occlusions robust facial expression recognition models, like region attention network (RAN)~\cite{wang2020region} and CNN with attention mechanism (ACNN)~\cite{li2018occlusion}.

\subsection{Masked Face Age Classification}

Similar to face and facial expression recognition systems,  age estimation techniques also critically depend on the visibility of the facial areas and struggle with performance when parts are occluded. As a result, studies investigating age estimation with facial masks have also appeared during the COVID-19 pandemic.

Golwalker et al.~\cite{golwalkar2020age} conjectured that using large prediction models in age estimation with occluded faces makes it challenging due to the lack of suitable large-scale datasets. When wearing masks, the most discriminative features for age estimation are largely hidden below the masks, like wrinkles on the cheeks and mouth. Their approach to this problem was, therefore, using a shallow model that could be fine-tuned easily based on a small set of images of people wearing masks. To this end, the authors used a simple 9-layer  CNN architecture. For the age detection dataset, they collected faces wearing masks from various age categories and augmented it with an auxiliary dataset of 4500 synthetic images of masked people using a Generative Adversarial Network (GAN)~\cite{gan}. Öztel et al.~\cite{yolcu2021multi} developed a two-stage pipeline consisting of a face mask detection and an age classification stage. With this approach, the authors first determine if the person is wearing a face mask or not. Then, depending on the result, two separate age classification models are utilized. If the person is not wearing a mask, a standard classification model in the form of a simple CNN trained on UTKFace Large Scale Face Dataset~\cite{zhang2017age} is used. If the person is wearing a face mask, another simple CNN model is utilized, but this time trained with simulated face masks on the UTKFace Large Scale Face Dataset~\cite{zhang2017age}. The proposed pipeline included three different age classes, i.e., teenager (12-20), middle-aged (21-64), and elderly (65+), and was shown to ensure competitive results.

\subsection{Landmark Localization and Alignment}
Facial landmark localization and alignment are essential components of various face-related applications, such as face recognition, facial pose estimation, 3D face reconstruction, emotion recognition, face synthesis, and face morphing, and falls into the category of supporting techniques given our taxonomy from Section~\ref{Sec: Background}. The main goal of landmark localization is to locate key points of the given 2D face image, such as the nose tip, eyebrow curve, mouth corners, eye centers, or eye corners among others. Until the pandemic, many successful facial landmark localization approaches have been developed by using thousands of annotated face images~\cite{dlib09,deng2020retinaface,Honari_2018_CVPR}. However, the widespread usage of face masks to prevent virus transmission has brought new challenges to landmark localization and alignment similarly as to many other face-based algorithms. As it was not possible to collect and label a new dataset for the task, most methods prefer to use existing datasets, especially the JD-landmark dataset~\cite{xiang20213rd} and place virtual facial masks on the face images. This is because labeling facial landmark points on images with facial masks is hard, especially for 68 or 106 points of landmarks, which is the most commonly used markup in the literature. The authors of~\cite{sha2021efficient} propose MaskFan, which is a lightweight convolutional neural network that uses depthwise separable convolutions and group operations. They also propose a novel loss function named Enhanced Wing loss, which gives less importance to errors made near facial masks. Facial landmark localization methods generally adopt L1 or L2 loss functions that focus on more considerable errors. Since predicting facial landmarks over facial masks is a hard task due to invisible parts of the face, applying L1 or L2 loss forces the model to pay more attention to large errors that most heavily impact performance. In~\cite{wen2021towards}, Wen et al. also propose a new architecture for the masked facial landmark localization problem. Their model consists of three different neural networks designed for: alignment, estimation, and refinement. They use downscaled face images in the alignment network and then align faces according to the reference pose. Then, the estimation network predicts 106 facial landmarks. Finally, their refinement model takes the non-masked region of the face, which is eyes and eyebrows and tries to generate more accurate predictions. Hu et al.~\cite{hu2021robust} adopt multi-knowledge distillation and a pose-aware resampling strategy. They aim to increase data diversity by sampling images with different face poses. 

All the presented works generate virtual masks and apply them on the JD-landmark dataset~\cite{xiang20213rd}, while studies with real masks are still largely missing from the literature due to the obvious ground-truth issues associated with such work. Suggestions for evaluation strategies that only consider visible landmarks have also been made in prior publications~\cite{xiang20213rd}. 

\begin{figure}[t!]
\subfloat{\includegraphics[width = 0.47\linewidth]{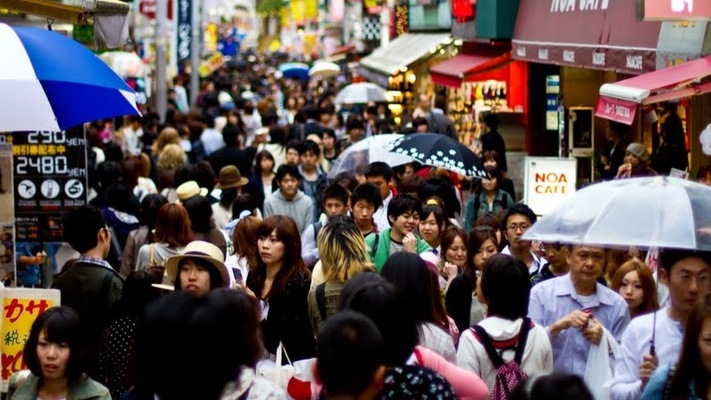}}\hfill
\subfloat{\includegraphics[width = 0.47\linewidth]{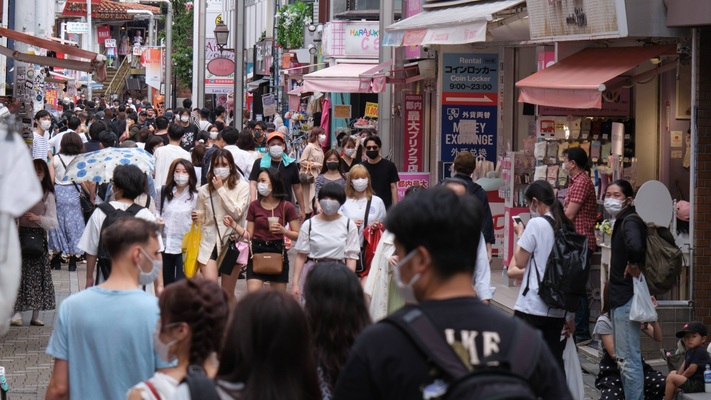}}\\
\caption{Generic images of crowded street before and during the pandemic, respectively. (\copyright\ JordyMeow and Benzoyl)}
\label{fig:masked_crowds}
\end{figure}

\begin{table}[tb]
\newcommand{\minitab}[2][l]{\begin{tabular}{#1}#2\end{tabular}}
\renewcommand{\arraystretch}{0.75}
\caption{Summary of the reviewed crowd counting methods}\label{tab:libs}
\smallskip
\centering
\resizebox{0.7\linewidth}{!}{%
\begin{tabular}{lcccc}
\toprule
\multirow{2}*{\minitab[@{}l@{}]{Method}} & \multirow{2}*{\minitab[@{}l@{}]{Year}} & \multirow{2}*{\minitab[@{}l@{}]{Handling of\\facial masks}} & \multirow{2}*{\minitab[@{}l@{}]{Handling of\\social distancing}} & \multirow{2}*{\minitab[@{}l@{}]{Mode of counting\\implementation}}\\\\
\midrule
\multirow{2}*{\minitab[@{}l@{}]{Al-Sa'd et al.~\cite{Alsad22}}} & \multirow{2}*{\minitab[@{}c@{}]{2022}} & \multirow{2}*{\minitab[@{}c@{}]{No}} & \multirow{2}*{\minitab[@{}c@{}]{Yes}} & \multirow{2}*{\minitab[@{}c@{}]{Detection based CNN}} \\\\
\multirow{2}*{\minitab[@{}l@{}]{Valencia et al.~\cite{Valencia21}}} & \multirow{2}*{\minitab[@{}c@{}]{2021}} & \multirow{2}*{\minitab[@{}c@{}]{No}} & \multirow{2}*{\minitab[@{}c@{}]{Yes}} & \multirow{2}*{\minitab[@{}c@{}]{Detection based CNN}}\\\\
\multirow{2}*{\minitab[@{}l@{}]{Somaldo et al.~\cite{Somaldo20}}} & \multirow{2}*{\minitab[@{}c@{}]{2020}} & \multirow{2}*{\minitab[@{}c@{}]{No}} & \multirow{2}*{\minitab[@{}c@{}]{Yes}} & \multirow{2}*{\minitab[@{}c@{}]{Detection based CNN}}\\\\
\multirow{2}*{\minitab[@{}l@{}]{Nguyen et al.~\cite{Nguyen21}}} & \multirow{2}*{\minitab[@{}c@{}]{2021}} & \multirow{2}*{\minitab[@{}c@{}]{No}} & \multirow{2}*{\minitab[@{}c@{}]{No}} & \multirow{2}*{\minitab[@{}c@{}]{Regression based CNN}} \\\\
\multirow{2}*{\minitab[@{}l@{}]{Almalki et al.~\cite{Almalki21}}} & \multirow{2}*{\minitab[@{}c@{}]{2021}} & \multirow{2}*{\minitab[@{}c@{}]{Yes}} & \multirow{2}*{\minitab[@{}c@{}]{No}} & \multirow{2}*{\minitab[@{}c@{}]{Detection based CNN}} \\\\
\multirow{2}*{\minitab[@{}l@{}]{He et al.~\cite{He22}}} & \multirow{2}*{\minitab[@{}c@{}]{2022 }} & \multirow{2}*{\minitab[@{}c@{}]{No}} & \multirow{2}*{\minitab[@{}c@{}]{No}} & \multirow{2}*{\minitab[@{}c@{}]{Attention based CNN}} \\\\
\multirow{2}*{\minitab[@{}l@{}]{Dosi et al.~\cite{dosi2021aecnet}}} & \multirow{2}*{\minitab[@{}c@{}]{2021}} & \multirow{2}*{\minitab[@{}c@{}]{No}} & \multirow{2}*{\minitab[@{}c@{}]{No}} & \multirow{2}*{\minitab[@{}c@{}]{Attention based CNN}} \\\\
\multirow{2}*{\minitab[@{}l@{}]{Alvarez et al.~\cite{Alvarez21}}} & \multirow{2}*{\minitab[@{}c@{}]{2021}} & \multirow{2}*{\minitab[@{}c@{}]{No}} & \multirow{2}*{\minitab[@{}c@{}]{Yes}} & \multirow{2}*{\minitab[@{}c@{}]{Detection based CNN}} \\\\
\multirow{2}*{\minitab[@{}l@{}]{Jarraya et al.~\cite{Kammoun21}}} & \multirow{2}*{\minitab[@{}c@{}]{2021}} & \multirow{2}*{\minitab[@{}c@{}]{No}} & \multirow{2}*{\minitab[@{}c@{}]{No}} & \multirow{2}*{\minitab[@{}c@{}]{Density based CNN}} \\\\
\multirow{2}*{\minitab[@{}l@{}]{Amin et al.~\cite{Amin21}}} & \multirow{2}*{\minitab[@{}c@{}]{2021}} & \multirow{2}*{\minitab[@{}c@{}]{Yes}} & \multirow{2}*{\minitab[@{}c@{}]{Yes}} & \multirow{2}*{\minitab[@{}c@{}]{Detection based CNN}} \\\\
\multirow{2}*{\minitab[@{}l@{}]{Nguyen et al.~\cite{nguyen2021effectiveness}}} & \multirow{2}*{\minitab[@{}c@{}]{2021}} & \multirow{2}*{\minitab[@{}c@{}]{Yes}} & \multirow{2}*{\minitab[@{}c@{}]{No}} & \multirow{2}*{\minitab[@{}c@{}]{Detection based CNN}} \\\\
\bottomrule
\end{tabular}}
\end{table}

\subsection{Crowd detection and counting}

One of the prevention measures to reduce the spread of the Coronavirus disease is physical/social distancing in public areas that can be monitored automatically using vision-based crowd counting techniques. Such techniques are able to count or estimate the number of people in a given area from a single image or a video acquired through surveillance cameras, CCTV or even drones. A plethora of research has been done over the past years on the crowd counting problem~\cite{Li21,Wang21a} dealing with challenges, such as mutual occlusions, non-uniform people density, varying scale, perspective, illumination, weather conditions, crowd size, and density, that can severely alter human appearance. In this section, we review crowd counting solutions with the focus on approaches developed to assist policy measures against the COVID-19 pandemic.

Early crowd counting methods mainly rely on object detection with counting. Usually, these methods first extract image features, such as shapelets~\cite{Sabzmeydani07}, Histograms of Oriented Gradients (HOGs)~\cite{Dalal05}, Haar wavelets~\cite{ShengFuu01} or other related descriptors from the image and then combine the computed representations with various classification methods, such as Support Vector Machines (SVMs)~\cite{Dalal05,ShengFuu01}, regression forests~\cite{Fiaschi12} and alike, in order to detect people in images. These methods work well for detecting sparse (masked) faces, but perform poorly on dense crowds where individual people are not clearly visible.

To alleviate the above-mentioned problems, some approaches rely on direct-count regression. Examples of such methods in~\cite{Ryan09} use handcrafted features and some regression technique, such as linear regression, to learn a mapping function between the features and the crowd count. These methods are able to accurately estimate people counts even in the presence of occlusions and background clutter, but they ignore spatial information. The solution to this problem is given by the density estimation based methods~\cite{Pham15} that learn a mapping between features in the local region and the corresponding object density maps, while integrating over the densities to obtain crowd counts. The approaches of this type generally use dot-annotated images for training that are transformed to density functions using kernel density estimation~\cite{Arteta14}.

With the advent of deep learning, many crowd counting approaches have been proposed based on convolutional neural networks~\cite{Fan2022}. 
Amin et al.~\cite{Amin21}, for example, proposed a solution that addresses both face mask detection and crowd counting. With this approach, the YOLO-based algorithm~\cite{Li19} is used to detect face masks, while the MobileNet single shot object detector~\cite{howard2017mobilenets} is used simultaneously for crowd counting. Kammoun-Jarraya et al.~\cite{Kammoun21} introduced a CNN based technique for crowd counting from a single image for enforcing social distancing during the COVID-19 pandemic. The proposed model follows the structure of VGG-19~\cite{vggmodel} with small kernel sizes in the convolutional layers but without the fully connected layers. Due to the fully convolutional structure, the model is able to process input images of arbitrary resolutions. The reported results on a new large-scale crowd counting dataset from the Saudi public areas point to the competitive performance compared to the state-of-art methods.
Alvarez et al.~\cite{Alvarez21} developed a software to monitor the physical distance and the crowd density of a specified area or a region of interest. The YOLOv3 model~\cite{yolov3} was used in this work to detect humans in each video frame captured by a mobile phone. Physical distancing is monitored through computing the interpersonal distance of a pair of centroids of the detected bounding boxes, while the crowd density is computed by counting the number of people present in the region of interest. The software is able to detect 83\% of physical distancing and 84\% of crowd density violations.
Dosi et al.~\cite{dosi2021aecnet} proposed a pipeline named Attentive EfficientNet (AECNet) for density estimation in crowd counting that makes use of an encoder-decoder-based architecture. In the encoder block, they use EfficientNet~\cite{efficientnetmodel} and empirically show their superiority over other feature extraction architectures.

To mitigate the problem of huge scale variations, He et al.~\cite{He22} proposed a novel approach for crowd counting named Jointly Attention Network (JANet). They designed the Multi-order Scale Attention module to extract meaningful high-order statistics with abundant scale details and also introduced the Multi-pooling Relational Channel Attention module to investigate the global scope relations and structural semantics. 
Various experiments illustrated the superiority of the JANet approach. Almalki et al.~\cite{Almalki21} introduced an approach that detects, counts, and classifies the crowd’s masking condition and calculates spatiotemporal safety index that can be used for assisting effective policy decisions and relief plans against COVID-19. The approach uses YOLOv3~\cite{yolov3} to extract image features and a classification layer is added at the end of the YOLOv3 extractor that classifies a detected face to either the mask or the no-mask group. 
A unified system that allows the scale variation problem to be solved both directly and indirectly was described by Nguyen et al. in~\cite{Nguyen21}. Here, the dense scale information is learned directly through the main network, which is designed with dense dilated convolution blocks and dense residual connections among the blocks. The scale information is further incorporated into the features indirectly through learning depth information from an auxiliary depth dataset.

Somaldo et al.~\cite{Somaldo20} proposed a drone that has the ability of localization, navigation, people detection, crowd identifier, and social distancing warning. For this purpose they utilize YOLOv3~\cite{yolov3} to detect people and also define an adaptive social distancing detector. 
Valencia et al.~\cite{Valencia21} presented a desktop application that utilizes YOLOv4-tiny~\footnote{\url{https://github.com/AlexeyAB/darknet}} and the DeepSORT tracking algorithm~\cite{deepsort} to monitor crowd counts and social distancing from a top-view camera perspective. 
A privacy-preserving adaptive social distance estimation and crowd monitoring solution for surveillance cameras was proposed by Al-Sa'd et al.~\cite{Alsad22}. The authors utilize OpenPose~\cite{Cao19} to detect and localize people. Their approach is able to compute inter-personal distances in real-world coordinates, detect social distance infractions and identify overcrowded regions in a scene. 
The work presented in~\cite{nguyen2021effectiveness} investigated the effectiveness of different approaches to estimate the ratio of people wearing a mask within an observed crowd - a problem referred to by the authors as \textit{mask-wearing ratio estimation}. Specifically, the authors compared detection-based and regression-based approaches to crowd counting, while also distinguishing between people with and without masks in the given crowd. Moreover, the authors improved the state-of-the-art face detector, RetinaFace~\cite{deng2020retinaface}, to be able to better estimate the mask-wearing ratio. A large-scale dataset with more than 580,000 face annotations was also introduced to facilitate the experiments.

\subsection{Breathing rate measurements}

Clinical studies on patients with COVID-19 disease showed that one of the most common symptoms are fever, respiratory and digestive symptoms~\cite{pan2020clinical}. In order to identify breathing abnormalities, which can be a symptom of COVID-19, multiple research works suggested measuring respiratory rate using wearable devices~\cite{natarajan2021measurement}, non-contact radar signals~\cite{non_contact_monitoring} and thermal cameras~\cite{queiroz2021thermal}.

Among these breathing rate measurement techniques, detecting breathing anomalies using thermal cameras represents a cheap and effective solution that can easily be implemented in practice, as many countries already deployed thermal cameras to detect people with high fever at airports and public buildings~\cite{perpetuini2021overview}. Following this line of research, Queiroz et al.~\cite{queiroz2021thermal} proposed to analyze the intensity of thermal images over time using deep learning techniques. The approach exploits the fact that the region covered by facial masks gets warmer when exhaling, which can be detected through the analysis of the pixel intensities in the thermal image. Similarly, when the pixel intensity within the  mask region decreases, this indicates that the person is inhaling. To facilitate the research, the authors collected $33$ videos of $11$ subjects, with subjects breathing slowly, normal and fast. Their experimental results showed that breathing rate measurements can reach an accuracy of up to 91\% on their dataset.

\subsection{Face-hand interaction detection}

\begin{figure}[t!]
     \centering
\includegraphics[width=0.9\linewidth]{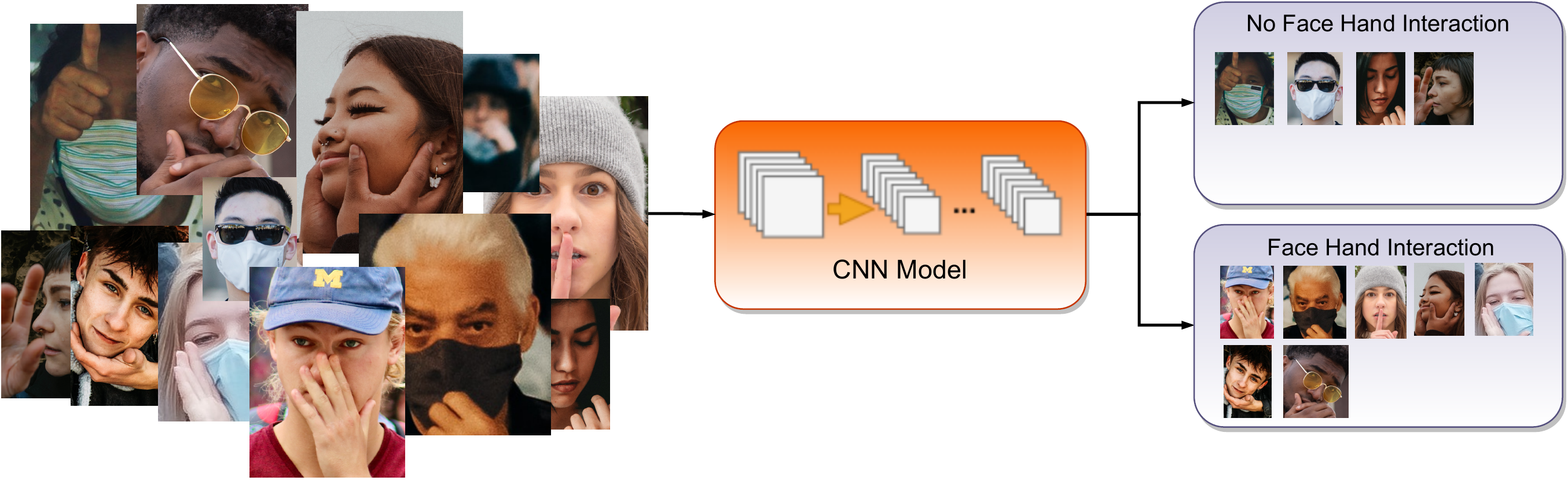}
\caption{Illustration of the face-hand interaction detection task. One of the most common advice given to prevent virus transmission is not to touch the face with hands. During the COVID-19 pandemic, automatic detection of face-hand interaction has gained importance as a research topic in CVHA.  }
\label{fig:face_hand_interaction}
\end{figure}

To minimize the transmission of COVID-19, common advice issued by health organizations included limiting face-hand interaction. CVHA techniques were also developed to help monitor face-hand interaction in public spaces.  A basic representation of the face-hand interaction detection task is presented in Figure \ref{fig:face_hand_interaction}.

One of the initial studies by Beyan et al.~\cite{beyan2020analysis} investigated the face-hand touching behavior of people. The authors first manually annotated $64$ video recordings, originally collected for the analysis of social interactions within a small group of people, for face-hand touching interaction. Next, they evaluated rule-based, hand-crafted feature-based, and learned CNN feature-based models for their performance in face-hand touching detection and found that the CNN model yielded the best overall results with an  F1-score of 83.76\%. In a more recent study, Eyiokur et al.~\cite{eyiokur2022unconstrained} explored the applicability of several well-known CNN models, such as ResNet~\cite{resnetmodel} and EfficientNet~\cite{efficientnetmodel}, for face-hand interaction detection. Here, the authors first introduced an unconstrained face-hand interaction dataset, named ISL-Unconstrained Face Hand Interaction Dataset (ISL-UFHD), to advance detection of face-hand interaction detection within a comprehensive prevention system for COVID-19 protection measurements, and then evaluated the considered classification models on the newly collected data. Experimental results showed that the highest classification accuracy of 93.35\% was obtained  with the EfficientNet-b2 model~\cite{efficientnetmodel}.

Both of the studies discussed above, proposed CVHA techniques that showed promise for the face-hand interaction problem. However, important challenges, such as the detection in extreme imaging conditions and under varying poses, or in the presence of ambiguity caused by the different depth levels of the face and hand, still persist. Face-hand interaction detection is, therefore, still considered an open research problem that requires further investigation. 

\subsection{Synthetic data generation and mask removal\label{SubSec:simulated}}

One of the main challenges in CVHA at the beginning of the COVID-19 was the obvious lack of suitable datasets needed to train various CVHA techniques. In response to this challenge, generative approaches have been quickly adopted to build synthetic datasets to alleviate the need of collecting real-life masked face images as well as to develop methods based on data augmentation and generation for various tasks such as face recognition, identification, and landmark detection.

To artificially generate face images with masks, Anwar et al.~\cite{anwar2020masked} developed an open-source tool, MaskTheFace\footnote{\url{https://github.com/aqeelanwar/MaskTheFace}}, that can convert non-masked faces to masked faces effectively. The tool uses the Dlib-based face landmark detector~\cite{dlib09} to identify the face tilt and six key features, i.e. landmarks, on the face to properly fit a face mask. Alternatively, Wang et al.~\cite{wang2021facex} presented another open-source toolbox, FaceX-Zoo\footnote{\url{https://github.com/JDAI-CV/FaceX-Zoo}}, which implements a Facial Mask Adding (FMA-3D)\footnote{\url{https://github.com/JDAI-CV/FaceX-Zoo/tree/main/addition_module/face_mask_adding/FMA-3D}} method for adding a mask to a non-masked face image. Given a real masked face image $I$ and a non-masked face image $J$, this method synthesizes a photo-realistic masked face image with the mask region coming from $I$ and the facial area coming from $J$.

Encouraged by these initiatives, many studies have attempted to enrich existing datasets containing faces without masks, e.g. CelebA~\cite{celebahq}, CASIA-WebFace, LFW~\cite{huang2008labeled}, CALFW~\cite{calfw}, with synthetically generated masked face images to enable further research on masked face recognition~\cite{wang2020masked,karasugi2020face,mare2021realistic,xiang20213rd,huang2021face,wang2021mlfw}. For instance, Wang et al.~\cite{wang2020masked} and Karasugi et al.~\cite{karasugi2020face} generated synthetic face mask datasets using Dlib's landmark detector~\cite{dlib09} to properly align face mask templates on faces, whereas Mare et al.~\cite{mare2021realistic} relied on SparkAR Studio\footnote{\url{https://sparkar.facebook.com/ar-studio/}}, a developer program made by Facebook to create Instagram face filters, to create synthetic masks and overlay them on faces in the original images. On the other hand, Xiang et al.~\cite{xiang20213rd} targeted to improve the accuracy and robustness of facial landmark localization on masked faces by introducing a new dataset with generated masks that are largely varied in identity, head pose, facial expression, and occlusion based on the FMA-3D method in the FaceX-Zoo toolbox~\cite{wang2021facex}.

Some studies tackled the opposite problem and focused on removing the face masks from images~\cite{din2020novel,li2021face,coelho2021generative,hu2021covertheface}. The idea with these studies is to bring the data closer to the real-world masked-free data, for which standard off-the shelf CVHA models for different tasks are readily available. For instance, Din et al.~\cite{din2020novel} investigated a two-stage method for unmasking of masked faces where the first stage detects and segments masks with a modified version of U-Net~\cite{unetmodel} and the second stage deploys a GAN-based network with global and local discriminators for mask-area inpainting. Similarly, Li et al.~\cite{li2021face} proposed a method combining a GAN and a texture network to first inpaint the face after removing the mask and then to smooth out the texture to make the resulting face more realistic. Taking a step further, Coelho et al.~\cite{coelho2021generative} presented a generative approach for face mask removal using audio and appearance together. This approach estimated landmarks representing mouth structure from the audio, and feed these landmarks into a GAN to reconstruct the full face image with the mouth in a correct shape. Hu et al.~\cite{hu2021covertheface} described a method to generate faces with properly worn masks, either by simply overlaying the mask if no mask is worn or by first removing and then overlaying the mask if the mask is incorrectly worn. The method employs Mean and Covariance Feature Matching GAN (MCGAN)~\cite{mcganmodel} for the mask removal task and uses MaskTheFace~\cite{anwar2020masked}.

\section{Datasets}\label{Sec: Datasets}

\begin{figure*}
    \centering
    \includegraphics[width=0.9\linewidth]{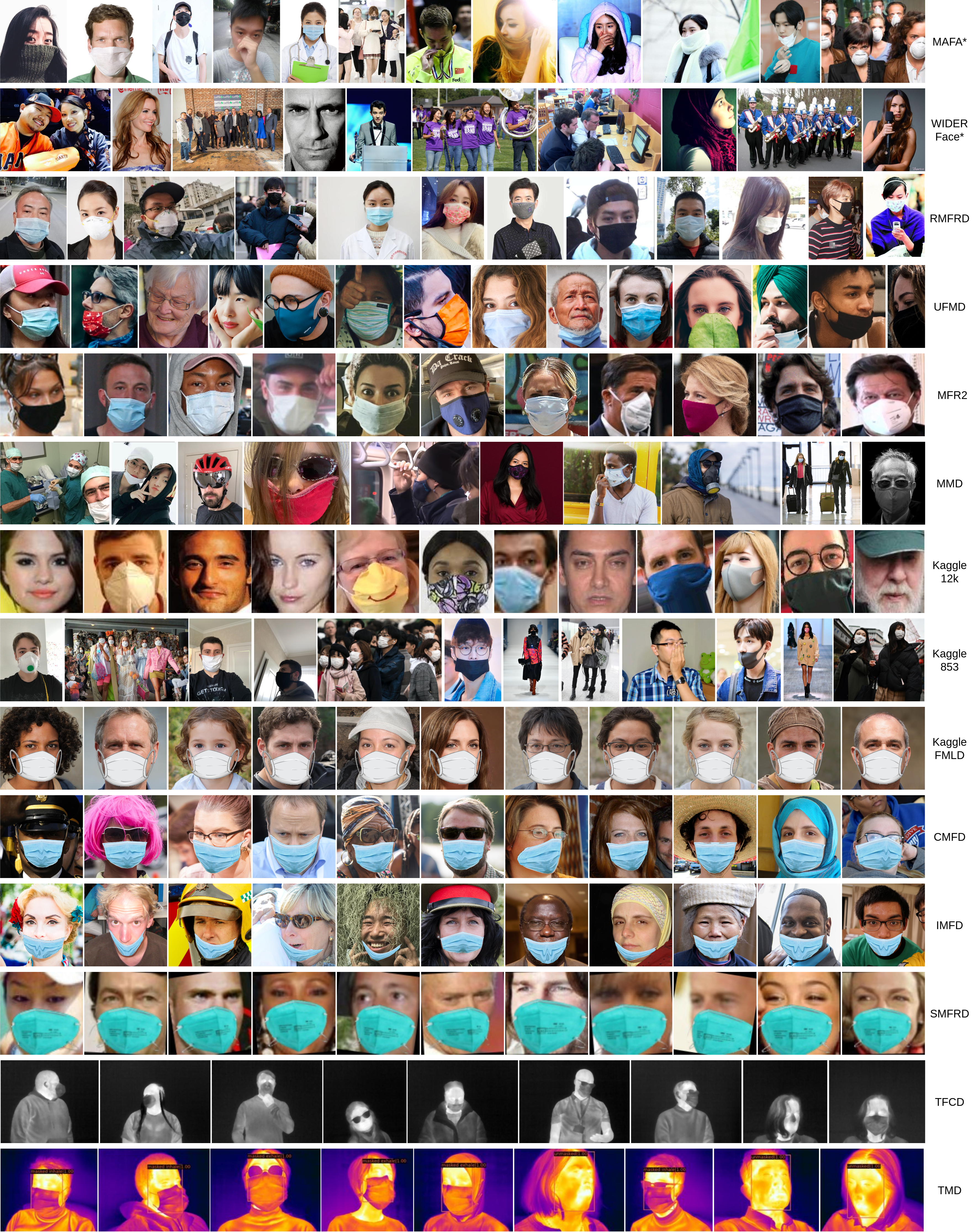}
    \caption{Illustrative example images from different datasets introduced for the development of CVHA techniques for the COVID-19 era.}
    \label{fig:dataset}
\end{figure*}

\begin{table}[thb!]
\centering
\caption{Summary of COVID-19 related datasets reviewed in this paper. BB: Face bounding box, FM: Face Mask, FP: Frontal to Profile, MF: Masked face, FO:Face Occlusion}
\newcommand{\minitab}[2][l]{\begin{tabular}{#1}#2\end{tabular}}
\renewcommand{\arraystretch}{1}
\label{tab:datasets}
\smallskip
\centering
\resizebox{\linewidth}{!}{%
\begin{tabular}{llcccccll}
\toprule
\multirow{2}*{\minitab[@{}l@{}]{Dataset name}} & 
\multirow{2}*{\minitab[@{}l@{}]{Dataset source \& availability}} &
\multirow{2}*{\minitab[@{}l@{}]{Mask Types}} &
\multirow{2}*{\minitab[@{}l@{}]{Number of images}} & 
\multirow{2}*{\minitab[@{}l@{}]{Head pose}} &
\multirow{2}*{\minitab[@{}l@{}]{Variation of subjects}} &
\multirow{2}*{\minitab[@{}l@{}]{Number of face\\mask classes}} &
\multirow{2}*{\minitab[@{}l@{}]{Purpose of\\data collection}} & 
\multirow{2}*{\minitab[@{}l@{}]{Annotation type}}\\\\
\midrule
\multirow{2}*{\minitab[@{}l@{}]{MAFA \cite{ge2017detecting}}} & 
\multirow{2}*{\minitab[@{}l@{}]{Internet\\Publicly available}} &
\multirow{2}*{\minitab[@{}l@{}]{Real}} &
\multirow{2}*{\minitab[@{}l@{}]{30,811}} & 
\multirow{2}*{\minitab[@{}l@{}]{Various}} &
\multirow{2}*{\minitab[@{}l@{}]{Medium Diversity\\Mostly Asian People}} &
\multirow{2}*{\minitab[@{}l@{}]{Multiple}} &
\multirow{2}*{\minitab[@{}l@{}]{MF detection}} & 
\multirow{2}*{\minitab[@{}l@{}]{BB coordinates\\FM \& occlusion classes}}\\\\

\multirow{2}*{\minitab[@{}l@{}]{MAFA-FMD \cite{jiang2020retinamask}}} & 
\multirow{2}*{\minitab[@{}l@{}]{MAFA\\Not publicly available}} &
\multirow{2}*{\minitab[@{}l@{}]{Real}} &
\multirow{2}*{\minitab[@{}l@{}]{56,084}} & 
\multirow{2}*{\minitab[@{}l@{}]{Various}} &
\multirow{2}*{\minitab[@{}l@{}]{Medium Diversity\\Mostly Asian People}} &
\multirow{2}*{\minitab[@{}l@{}]{3}} &
\multirow{2}*{\minitab[@{}l@{}]{FM detection}} & 
\multirow{2}*{\minitab[@{}l@{}]{BB coordinates\\FM classes}}\\\\

\multirow{2}*{\minitab[@{}l@{}]{FMLD \cite{batagelj2021correctly}}} & 
\multirow{2}*{\minitab[@{}l@{}]{MAFA,Wider Face\\Publicly available}} &
\multirow{2}*{\minitab[@{}l@{}]{Real}} &
\multirow{2}*{\minitab[@{}l@{}]{63,072}} & 
\multirow{2}*{\minitab[@{}l@{}]{Various}} &
\multirow{2}*{\minitab[@{}l@{}]{High Diversity}} &
\multirow{2}*{\minitab[@{}l@{}]{3}} &
\multirow{2}*{\minitab[@{}l@{}]{MF detection}} & 
\multirow{2}*{\minitab[@{}l@{}]{BB coordinates\\FM \& occlusion classes}}\\\\

\multirow{2}*{\minitab[@{}l@{}]{WearMask \cite{wang2021wearmask}}} & 
\multirow{2}*{\minitab[@{}l@{}]{MAFA,Wider Face,Internet\\Publicly available}} &
\multirow{2}*{\minitab[@{}l@{}]{Real}} &
\multirow{2}*{\minitab[@{}l@{}]{9,097}} & 
\multirow{2}*{\minitab[@{}l@{}]{Various}} &
\multirow{2}*{\minitab[@{}l@{}]{High Diversity}} &
\multirow{2}*{\minitab[@{}l@{}]{2}} &
\multirow{2}*{\minitab[@{}l@{}]{MF detection}} & 
\multirow{2}*{\minitab[@{}l@{}]{BB coordinates\\FM classes}}\\\\

\multirow{2}*{\minitab[@{}l@{}]{PWMFD \cite{jiang2021real}}} & 
\multirow{2}*{\minitab[@{}l@{}]{MAFA,Wider Face,Internet\\Publicly available}} &
\multirow{2}*{\minitab[@{}l@{}]{Real}} &
\multirow{2}*{\minitab[@{}l@{}]{9,205}} & 
\multirow{2}*{\minitab[@{}l@{}]{Various}} &
\multirow{2}*{\minitab[@{}l@{}]{High Diversity}} &
\multirow{2}*{\minitab[@{}l@{}]{3}} &
\multirow{2}*{\minitab[@{}l@{}]{MF detection}} & 
\multirow{2}*{\minitab[@{}l@{}]{BB coordinates\\FM classes}}\\\\

\multirow{2}*{\minitab[@{}l@{}]{ISL-UFMD \cite{eyiokur2022unconstrained}}} & 
\multirow{2}*{\minitab[@{}l@{}]{Internet,Celebahq, FFHQ\\Publicly available}} &
\multirow{2}*{\minitab[@{}l@{}]{Real}} &
\multirow{2}*{\minitab[@{}l@{}]{20,891}} & 
\multirow{2}*{\minitab[@{}l@{}]{Various}} &
\multirow{2}*{\minitab[@{}l@{}]{High Diversity}} &
\multirow{2}*{\minitab[@{}l@{}]{3}} &
\multirow{2}*{\minitab[@{}l@{}]{FM detection}} & 
\multirow{2}*{\minitab[@{}l@{}]{FM classes}}\\\\

\multirow{2}*{\minitab[@{}l@{}]{SMFRD \cite{wang2020masked}}} & 
\multirow{2}*{\minitab[@{}l@{}]{Face recg. Datasets\\Publicly available}} &
\multirow{2}*{\minitab[@{}l@{}]{Artificial}} &
\multirow{2}*{\minitab[@{}l@{}]{500,000}} & 
\multirow{2}*{\minitab[@{}l@{}]{Various}} &
\multirow{2}*{\minitab[@{}l@{}]{Various}} &
\multirow{2}*{\minitab[@{}l@{}]{2}} &
\multirow{2}*{\minitab[@{}l@{}]{MF recognition}} & 
\multirow{2}*{\minitab[@{}l@{}]{FM classes\\Subject ids}}\\\\

\multirow{2}*{\minitab[@{}l@{}]{MaskedFace-Net\\CMFD,IMFD \cite{cabani2021maskedface}}} & 
\multirow{2}*{\minitab[@{}l@{}]{FFHQ\\Publicly available}} &
\multirow{2}*{\minitab[@{}l@{}]{Artificial}} &
\multirow{2}*{\minitab[@{}l@{}]{133,783}} & 
\multirow{2}*{\minitab[@{}l@{}]{Mostly\\ Frontal}} &
\multirow{2}*{\minitab[@{}l@{}]{High Diversity}} &
\multirow{2}*{\minitab[@{}l@{}]{2}} &
\multirow{2}*{\minitab[@{}l@{}]{FM detection}} & 
\multirow{2}*{\minitab[@{}l@{}]{FM classes\\Incorrect FM sub-classes}}\\\\

\multirow{2}*{\minitab[@{}l@{}]{MS1MV2-Masked \cite{boutros2022self}}} & 
\multirow{2}*{\minitab[@{}l@{}]{MS1MV2\\Publicly available}} &
\multirow{2}*{\minitab[@{}l@{}]{Artificial}} &
\multirow{2}*{\minitab[@{}l@{}]{57.5m}} & 
\multirow{2}*{\minitab[@{}l@{}]{Various}} &
\multirow{2}*{\minitab[@{}l@{}]{High Diversity}} &
\multirow{2}*{\minitab[@{}l@{}]{6}} &
\multirow{2}*{\minitab[@{}l@{}]{MF recognition}} & 
\multirow{2}*{\minitab[@{}l@{}]{FM classes\\Subject ids}}\\\\

\multirow{2}*{\minitab[@{}l@{}]{RMFRD \cite{wang2020masked}}} & 
\multirow{2}*{\minitab[@{}l@{}]{Internet\\Publicly available}} &
\multirow{2}*{\minitab[@{}l@{}]{Real}} &
\multirow{2}*{\minitab[@{}l@{}]{92,671}} & 
\multirow{2}*{\minitab[@{}l@{}]{FP}} &
\multirow{2}*{\minitab[@{}l@{}]{Low Diversity\\Asian People}} &
\multirow{2}*{\minitab[@{}l@{}]{2}} &
\multirow{2}*{\minitab[@{}l@{}]{MF recognition}} & 
\multirow{2}*{\minitab[@{}l@{}]{FM classes\\Subject ids}}\\\\

\multirow{2}*{\minitab[@{}l@{}]{Damer's MFRD \cite{damer2020effect}}} & 
\multirow{2}*{\minitab[@{}l@{}]{Collected by researchers\\Not publicly available}} &
\multirow{2}*{\minitab[@{}l@{}]{Real}} &
\multirow{2}*{\minitab[@{}l@{}]{2,160}} & 
\multirow{2}*{\minitab[@{}l@{}]{FP}} &
\multirow{2}*{\minitab[@{}l@{}]{Low Diversity}} &
\multirow{2}*{\minitab[@{}l@{}]{2}} &
\multirow{2}*{\minitab[@{}l@{}]{MF recognition}} & 
\multirow{2}*{\minitab[@{}l@{}]{Subject ids}}\\\\

\multirow{2}*{\minitab[@{}l@{}]{Damer's\\ MFRD-Extended \cite{damer2021extended}}} & 
\multirow{2}*{\minitab[@{}l@{}]{Collected by researchers\\Not publicly available}} &
\multirow{2}*{\minitab[@{}l@{}]{Real,\\Artificial}} &
\multirow{2}*{\minitab[@{}l@{}]{8,640}} & 
\multirow{2}*{\minitab[@{}l@{}]{FP}} &
\multirow{2}*{\minitab[@{}l@{}]{Low Diversity}} &
\multirow{2}*{\minitab[@{}l@{}]{2}} &
\multirow{2}*{\minitab[@{}l@{}]{MF recognition}} & 
\multirow{2}*{\minitab[@{}l@{}]{Subject ids}}\\\\

\multirow{2}*{\minitab[@{}l@{}]{MFR2 \cite{anwar2020masked}}} & 
\multirow{2}*{\minitab[@{}l@{}]{Internet\\Publicly available}} &
\multirow{2}*{\minitab[@{}l@{}]{Real}} &
\multirow{2}*{\minitab[@{}l@{}]{269}} & 
\multirow{2}*{\minitab[@{}l@{}]{FP}} &
\multirow{2}*{\minitab[@{}l@{}]{High Diversity}} &
\multirow{2}*{\minitab[@{}l@{}]{Multiple}} &
\multirow{2}*{\minitab[@{}l@{}]{MF recognition}} & 
\multirow{2}*{\minitab[@{}l@{}]{FM classes\\Subject ids}}\\\\

\multirow{2}*{\minitab[@{}l@{}]{FaceMask \cite{vrigkas2022facemask}}} & 
\multirow{2}*{\minitab[@{}l@{}]{Internet\\Publicly available}} &
\multirow{2}*{\minitab[@{}l@{}]{Real}} &
\multirow{2}*{\minitab[@{}l@{}]{4,866}} & 
\multirow{2}*{\minitab[@{}l@{}]{Various}} &
\multirow{2}*{\minitab[@{}l@{}]{Various}} &
\multirow{2}*{\minitab[@{}l@{}]{2}} &
\multirow{2}*{\minitab[@{}l@{}]{MF detection}} & 
\multirow{2}*{\minitab[@{}l@{}]{BB coordinates\\FM classes}}\\\\

\multirow{2}*{\minitab[@{}l@{}]{DS-IMF \cite{9667057}}} & 
\multirow{2}*{\minitab[@{}l@{}]{Collected by researchers\\Not publicly available}} &
\multirow{2}*{\minitab[@{}l@{}]{Real}} &
\multirow{2}*{\minitab[@{}l@{}]{3600}} & 
\multirow{2}*{\minitab[@{}l@{}]{Frontal}} &
\multirow{2}*{\minitab[@{}l@{}]{Low Diversity\\Indian People}} &
\multirow{2}*{\minitab[@{}l@{}]{2}} &
\multirow{2}*{\minitab[@{}l@{}]{MF detection\\ \& recognition}} & 
\multirow{2}*{\minitab[@{}l@{}]{BB coordinates\\Subject ids}}\\\\

\multirow{2}*{\minitab[@{}l@{}]{Thermal-Mask \cite{queiroz2021thermal}}} & 
\multirow{2}*{\minitab[@{}l@{}]{SpeakingFaces\\Not publicly available}} &
\multirow{2}*{\minitab[@{}l@{}]{Artificial,\\Thermal}} &
\multirow{2}*{\minitab[@{}l@{}]{75,908}} & 
\multirow{2}*{\minitab[@{}l@{}]{FP}} &
\multirow{2}*{\minitab[@{}l@{}]{Low Diversity}} &
\multirow{2}*{\minitab[@{}l@{}]{2}} &
\multirow{2}*{\minitab[@{}l@{}]{Thermal MF detection\\Breathing rate measurement}} & 
\multirow{2}*{\minitab[@{}l@{}]{BB coordinates\\FM classes}}\\\\

\multirow{2}*{\minitab[@{}l@{}]{Thermal mask \\dataset \cite{glowacka2021face}}} & 
\multirow{2}*{\minitab[@{}l@{}]{Collected by researchers\\Not publicly available}} &
\multirow{2}*{\minitab[@{}l@{}]{Real,\\Thermal}} &
\multirow{2}*{\minitab[@{}l@{}]{7,920}} & 
\multirow{2}*{\minitab[@{}l@{}]{Various}} &
\multirow{2}*{\minitab[@{}l@{}]{Low Diversity}} &
\multirow{2}*{\minitab[@{}l@{}]{2}} &
\multirow{2}*{\minitab[@{}l@{}]{Thermal MF detection}} & 
\multirow{2}*{\minitab[@{}l@{}]{BB coordinates\\FM classes}}\\\\

\multirow{2}*{\minitab[@{}l@{}]{COVID-19 TFCD \cite{ward2021flunet}}} & 
\multirow{2}*{\minitab[@{}l@{}]{Collected by researchers\\Publicly available}} &
\multirow{2}*{\minitab[@{}l@{}]{Real,\\Thermal}} &
\multirow{2}*{\minitab[@{}l@{}]{261}} & 
\multirow{2}*{\minitab[@{}l@{}]{FP}} &
\multirow{2}*{\minitab[@{}l@{}]{Low Diversity}} &
\multirow{2}*{\minitab[@{}l@{}]{2}} &
\multirow{2}*{\minitab[@{}l@{}]{Thermal MF detection}} & 
\multirow{2}*{\minitab[@{}l@{}]{BB coordinates\\FM classes}}\\\\

\multirow{2}*{\minitab[@{}l@{}]{Kaggle853 \cite{kaggle853}}} & 
\multirow{2}*{\minitab[@{}l@{}]{Internet\\Publicly available}} &
\multirow{2}*{\minitab[@{}l@{}]{Real}} &
\multirow{2}*{\minitab[@{}l@{}]{853}} & 
\multirow{2}*{\minitab[@{}l@{}]{Various}} &
\multirow{2}*{\minitab[@{}l@{}]{Medium Diversity\\Mostly Asian}} &
\multirow{2}*{\minitab[@{}l@{}]{3}} &
\multirow{2}*{\minitab[@{}l@{}]{MF detection}} & 
\multirow{2}*{\minitab[@{}l@{}]{BB coordinates\\FM classes}}\\\\

\multirow{2}*{\minitab[@{}l@{}]{Kaggle12k \cite{kaggle12k}}} & 
\multirow{2}*{\minitab[@{}l@{}]{Internet\\Publicly available}} &
\multirow{2}*{\minitab[@{}l@{}]{Real}} &
\multirow{2}*{\minitab[@{}l@{}]{~12,000}} & 
\multirow{2}*{\minitab[@{}l@{}]{FP}} &
\multirow{2}*{\minitab[@{}l@{}]{High Diversity}} &
\multirow{2}*{\minitab[@{}l@{}]{2}} &
\multirow{2}*{\minitab[@{}l@{}]{FM detection}} & 
\multirow{2}*{\minitab[@{}l@{}]{FM classes}}\\\\

\multirow{2}*{\minitab[@{}l@{}]{Kaggle\\FMLD \cite{kagglefml}}} & 
\multirow{2}*{\minitab[@{}l@{}]{Style Gan-2 Generated\\Publicly available}} &
\multirow{2}*{\minitab[@{}l@{}]{Artificial}} &
\multirow{2}*{\minitab[@{}l@{}]{20,000}} & 
\multirow{2}*{\minitab[@{}l@{}]{Frontal}} &
\multirow{2}*{\minitab[@{}l@{}]{High Diversity}} &
\multirow{2}*{\minitab[@{}l@{}]{2}} &
\multirow{2}*{\minitab[@{}l@{}]{Not clarified}} & 
\multirow{2}*{\minitab[@{}l@{}]{FM classes}}\\\\

\multirow{2}*{\minitab[@{}l@{}]{WWMR-DB \cite{wwwmrdb2021}}} & 
\multirow{2}*{\minitab[@{}l@{}]{Collected by researchers\\Publicly available}} &
\multirow{2}*{\minitab[@{}l@{}]{Real}} &
\multirow{2}*{\minitab[@{}l@{}]{1222}} & 
\multirow{2}*{\minitab[@{}l@{}]{FP}} &
\multirow{2}*{\minitab[@{}l@{}]{Low Diversity}} &
\multirow{2}*{\minitab[@{}l@{}]{Multiple}} &
\multirow{2}*{\minitab[@{}l@{}]{MF detection}} & 
\multirow{2}*{\minitab[@{}l@{}]{BB coordinates\\FM classes}}\\\\

\multirow{2}*{\minitab[@{}l@{}]{Medical Mask Dataset\\(MMD) \cite{mmd}}} & 
\multirow{2}*{\minitab[@{}l@{}]{Public domain\\Publicly available}} &
\multirow{2}*{\minitab[@{}l@{}]{Real}} &
\multirow{2}*{\minitab[@{}l@{}]{~6,000}} & 
\multirow{2}*{\minitab[@{}l@{}]{Various}} &
\multirow{2}*{\minitab[@{}l@{}]{High Diversity}} &
\multirow{2}*{\minitab[@{}l@{}]{3}} &
\multirow{2}*{\minitab[@{}l@{}]{MF detection}} & 
\multirow{2}*{\minitab[@{}l@{}]{BB coordinates\\FM \& occlusion classes}}\\\\

\multirow{2}*{\minitab[@{}l@{}]{AIZOOTech face \\mask dataset \cite{AIZOOTech}}} & 
\multirow{2}*{\minitab[@{}l@{}]{MAFA,Wider Face\\Publicly available}} &
\multirow{2}*{\minitab[@{}l@{}]{Real}} &
\multirow{2}*{\minitab[@{}l@{}]{7,971}} & 
\multirow{2}*{\minitab[@{}l@{}]{Various}} &
\multirow{2}*{\minitab[@{}l@{}]{High Diversity}} &
\multirow{2}*{\minitab[@{}l@{}]{2}} &
\multirow{2}*{\minitab[@{}l@{}]{MF detection}} & 
\multirow{2}*{\minitab[@{}l@{}]{BB coordinates\\FM classes}}\\\\

\multirow{2}*{\minitab[@{}l@{}]{BAFMD \cite{biasaware2022kantarci}}} & 
\multirow{2}*{\minitab[@{}l@{}]{Twitter\\Publicly available}} &
\multirow{2}*{\minitab[@{}l@{}]{Real}} &
\multirow{2}*{\minitab[@{}l@{}]{6,264}} & 
\multirow{2}*{\minitab[@{}l@{}]{Various}} &
\multirow{2}*{\minitab[@{}l@{}]{High Diversity}} &
\multirow{2}*{\minitab[@{}l@{}]{2}} &
\multirow{2}*{\minitab[@{}l@{}]{MF detection}} & 
\multirow{2}*{\minitab[@{}l@{}]{BB coordinates\\FM classes}}\\\\

\multirow{2}*{\minitab[@{}l@{}]{ROF \cite{erakin2021occl}}} & 
\multirow{2}*{\minitab[@{}l@{}]{Google Image Search\\Publicly available}} &
\multirow{2}*{\minitab[@{}l@{}]{Real}} &
\multirow{2}*{\minitab[@{}l@{}]{5,559}} & 
\multirow{2}*{\minitab[@{}l@{}]{Various}} &
\multirow{2}*{\minitab[@{}l@{}]{High Diversity}} &
\multirow{2}*{\minitab[@{}l@{}]{3}} &
\multirow{2}*{\minitab[@{}l@{}]{MF recognition}} & 
\multirow{2}*{\minitab[@{}l@{}]{FO classes\\Subject ids}}\\\\

\bottomrule
\end{tabular}}
\end{table}

An unprecedented number of datasets targeting various CVHA tasks related to COVID-19 have been introduced over the last few years. While masked face datasets were already collected back in 2017~\cite{ge2017detecting}, the need for suitable larger-scale collections of masked face images increased significantly during the COVID-19 pandemic. As a  result, novel (real and simulated) datasets were introduced for masked face detection and recognition, face-hand interaction detection, (masked) crowd counting and as well as other related CVHA problems.
In Table~\ref{tab:datasets}, we summarize the main COVID-19 related datasets and compare their characteristics. 

In~\cite{ge2017detecting}, the first large-scale masked face dataset, named MAFA, was published. The dataset contains 30,811 images of multiple persons with various head poses, face occlusions, and ethnicity, collected from the Internet. The MAFA dataset includes 35,806 masked face crops (there are multiple faces per image) with six annotated attributes: face, eye, and mask coordinates, head pose, occlusion degree, and four different mask types.
The dataset is primarily intended for the development of face/mask detection models. However, it needs to be noted that some of the masks present in the data are not worn correctly, e.g., they are not covering the nose, so mask detection models developed on this dataset are generally considered less suitable for monitoring applications aimed at preventing the spread of the COVID-19 disease. 
To address this issue, some studies considered improper mask usage as an additional label for the facial images, i.e., next to the \textit{mask} and \textit{no mask} labels.
Such data labels helps conceive more appropriate systems with respect to health-protective rules and usability in real-world conditions. In~\cite{batagelj2021correctly}, Batagelj et al. pointed out that the actual annotations of MAFA~\cite{ge2017detecting} are not suitable for training useful detectors to distinguish between correctly and incorrectly worn masks. The authors, therefore, reannotated the MAFA images based on health-protective rules. In addition to MAFA~\cite{ge2017detecting}, they also annotated the Wider Face dataset and released the generated annotations under the name Face Mask Label Dataset (FMLD). Thus, FMLD~\cite{batagelj2021correctly} includes images partitioned into three groups: 29,532 images with correctly worn masks, 1,528 images with incorrectly worn masks, and 32,012 images with mask free faces. In addition to mask annotations, the FMLD also has bounding box coordinates of faces, gender, ethnicity, and pose labels for each face. 
Wang et al.~\cite{wang2021wearmask} utilized the same benchmark datasets, Wider Face~\cite{yang2016wider} \& MAFA~\cite{ge2017detecting}, to build a serverless edge face detection tool. Their dataset included 4,065 images from MAFA, 3,894 images from Wider Face~\cite{yang2016wider}, and 1,138 additional images from the Internet, for a total of 17,532 face crops with corresponding bounding boxes. In~\cite{jiang2021real}, a new dataset called the Properly Wearing Masked Face Detection Dataset (PWMFD) is presented and consists of 3,615 newly collected images, 2,581 relabeled images from MAFA~\cite{ge2017detecting}, 2,951 images from Wider Face~\cite{yang2016wider}, and 58 images from RMFRD ~\cite{wang2020masked}. Similar to the previous studies, Jiang et al.~\cite{jiang2021real} considered three classes for the labels of their dataset, i.e., correctly worn, incorrectly worn and mask-free.
In total, there are 7,695 properly worn masked faces, 10,471 mask-free faces, and 366 incorrectly worn masked faces in PWMFD. In~\cite{jiang2020retinamask}, a face detector was first applied to the MAFA dataset~\cite{ge2017detecting}, and the generated face crops were then reannotated with respect to virus protection rules. This way, a new dataset, named MAFA-FMD, was collected and includes 56,024 images belonging to correct, incorrect, and no mask-wearing classes. Unfortunately, the MAFA-FMD dataset is not publicly available.  

Eyiokur et al.~\cite{eyiokur2022unconstrained} proposed an unconstrained masked face dataset~\footnote{\url{https://github.com/iremeyiokur/COVID-19-Preventions-Control-System}}, named ISL-Unconstrained Face Mask Dataset (ISL-UFMD), to study CVHA techniques for COVID-19.  ISL-UFMD~\cite{eyiokur2022unconstrained} contains 11,075 mask-free, 9,300 proper, and 513 improper mask images, collected from the Internet, YouTube videos, and well-known face datasets, such as CelebA-HQ~\cite{celebahq} and LFW~\cite{huang2008labeled}. By relying on different resources during data collection, the data in ISL-UFMD features highly diverse images captured in unconstrained conditions with variability across ethnicity, age, gender, head pose, and  environmental settings. Furthermore, Eyiokur et al.~\cite{eyiokur2022unconstrained} also presented the first unconstrained face-hand interaction dataset named ISL-Unconstrained Face Hand Interaction Dataset (ISL-UFHD) to advance face-hand interaction detection with respect to COVID-19 protection rules. In ISL-UFHD, there are 10,018 samples with face-hand interaction and 20,038 without. Another related dataset, FaceMask, was described by Vrinkas et al. in~\cite{vrigkas2022facemask} and contains 4,866 images of people with variations in gender and ethnicity, occlusions, and capture conditions, e.g., indoor/outdoor. Some of the faces are blurred, have partial occlusions or are of low-resolution due to distance to the camera. There are 15,419 and 12,262 face crops that belong to the \textit{mask} and \textit{no mask} classes, respectively. Morever, Kantarci et al.~\cite{biasaware2022kantarci} proposed Bias Aware Face Mask Detection (BAFMD) dataset in order to create a dataset that minimizes potential bias on ethnicity, age, and gender. Their dataset contains 6,264 images from Twitter with more than 16,000 facial bounding boxes with and without facial masks.

The need for a large-scale masked face datasets motivated researchers to also generate simulated images with artificial masks positioned on the face, as already discussed in Section~\ref{SubSec:simulated}. In~\cite{cabani2021maskedface}, a large-scale simulated masked face dataset named MaskedFace-Net, which includes the Correctly worn Masked Face Dataset (CMFD) and Incorrectly worn Masked Face Dataset (IMFD) subsets, was presented. MaskedFace-Net was constructed from the Flickr-Faces-HQ3 (FFHQ) dataset~\cite{karras2019style} using a mask-to-face deformable model and contains 137,016 images in total.  
In~\cite{huber2021mask}, the popular large-scale MS1MV2 dataset~\cite{guo2016ms} with 5.8M images of 85k subjects was augmented with simulated face masks with a probability of 0.5. Similarly, in~\cite{boutros2022self}, a face mask simulated version of the MS1MV2 dataset~\cite{guo2016ms} was utilized for the training of the presented masked face recognition system. To evaluate the proposed system, other well-known benchmarks for face verification, namely  IARPA Janus Benchmark-C (IJB-C) dataset~\cite{maze2018iarpa} and LFW~\cite{huang2008labeled}, were used to generate face images with synthetic face masks. Moreover, in~\cite{wang2020masked}, three novel datasets, named Masked Face Detection Dataset (MFDD), Real-world Masked Face Recognition Dataset (RMFRD), and Simulated Masked Face Recognition Dataset (SMFRD) were published to investigate the face mask detection and face recognition performance in the case of occlusion due to face masks. Wang et al.~\cite{wang2020masked} proposed 500,000 simulated masked face images of 10,000 subjects constructed with an artificial mask generation tool.  

In addition to the works that investigate prevention, monitoring and control CVHA techniques, e.g., for the detection and tracking of proper usage of face masks, some studies also examined the effect of wearing face masks on the performance of face recognition systems. To facilitate this work, novel real-world and simulated masked face recognition datasets were introduced. Wang et al.~\cite{wang2020masked}, for example, described the Real-world Masked Face Recognition Dataset (RMFRD) which consists of 5,000 masked and 90,000 non-masked face images that belong to 525 celebrities. 
Although it is stated that the RMFRD dataset contains 5,000 face images with masks, there are only 2,203 face images with masks in the publicly available version. 
Damer et al.~\cite{damer2020effect} presented a database that consists of 2,160 images of 24 participants from three different sessions. For each session, three videos are collected from the participants; two of them containing faces with and without a face mask in daylight and the third one containing faces with masks in different lighting condition. Session one was considered as a reference, and sessions two and three were considered as sources for the probe data. 
In their follow-up work~\cite{damer2021extended}, the same authors extended the initial dataset with an additional 24 participants and a new type of face images with simulated masks. 
In~\cite{anwar2020masked}, the authors published a relatively small dataset, called Masked Faces in Real World for Face Recognition (MFR2), that contains 53 identities with an average of five images, with or without face mask, per subject.

In~\cite{9667057}, Mishra et al. focused on analyzing masked face detection, gender prediction, mask/no mask classification, and masked face recognition on images that were acquired from Indian subjects. They introduced the Dual Sensor Indian Masked Face Dataset (DS-IMF), which consists of 300 subjects with 300 mask-free and 1,500 mask images per class. The images were captured with a DSLR camera and a mobile phone. Moreover, Fang et al.~\cite{FangPR_PAD} presented the novel Collaborative Real Mask Attack Database (CRMA) to investigate the effect of face masks on presentation attack detection. The CRMA dataset consists of 30\% AM0 (unmasked face PA), 60\% AM1 (masked face PA), and 10\% AM2 (unmasked face PA with a real masked placed on the PA) which are images of three different presentation attacks for analyzing both print and replay attacks. In~\cite{erakin2021occl}, a new real-world dataset named Real World Occluded Faces (ROF) with 3,195 neutral images, 1,686 sunglasses images (upper-face occluded) and 678 masked images (lower-face occluded)is presented. In ROF dataset, collected images belong to 180 different identities and they are used to explore effect of occlusions on face recognition performance.

Another notable group of works~\cite{queiroz2021thermal, glowacka2021face, ward2021flunet} focused on the collection of datasets for COVID-19 related applications using  thermal imaging. Queiroz et al.~\cite{queiroz2021thermal}, for example, utilized a large-scale multimodal dataset known as SpeakingFaces~\cite{abdrakhmanova2021speakingfaces} that consists of thermal images as well as visual and audio streams. The original SpeakingFaces dataset~\cite{abdrakhmanova2021speakingfaces} does not include faces with masks. Queiroz et al.~\cite{queiroz2021thermal}, therefore, generated thermal masked faces using artificial masks placed over the mouth and nose area. After preprocessing, 42,460 thermal masked and 33,448 thermal mask-free faces were included in the final Thermal-Mask Dataset (TMD)~\cite{queiroz2021thermal}.
In~\cite{glowacka2021face}, Glowacka et al. collected 7,920 thermal images with four different cameras from various distances and subjects with and without facial masks. The captured dataset~\cite{glowacka2021face} in the final form includes 10,555 faces, as some of the recorded images include multiple people. In~\cite{ward2021flunet}, a small thermal mask dataset (COVID-19 TFCD), with 250 images belonging to 20 participants, was collected. 

There are many online dataset repositories such as Kaggle, IEEE DataPort, and Github that allow researchers to publish their data collections. With growing interest in CVHA problems during the COVID-19 times, several face masked datasets~\cite{kaggle853, kaggle12k, kagglefml, wwwmrdb2021, mmd, AIZOOTech} were published on these repositories. In~\cite{kaggle853}, there are 853 images of 4,080 faces belonging to three face classes (present/not present/improperly worn). In~\cite{kaggle12k}, around 12k face images belonging to two main classes: mask, no mask, were published. The dataset has variations in terms of resolution, mask type, diverse people. Another dataset named Face Mask Lite Dataset (Kaggle-FMLD)~\cite{kagglefml} on Kaggle contains 10,000 artificial face images generated using StyleGAN2 architecture. By adding artificial masks to the generated faces, the authors  created a simulated dataset to address the masked face recognition problem. The Ways to Wear a Mask or a Respirator (WWMR-DB) dataset~\cite{wwwmrdb2021}, published on the IEEE DataPort, consists of 1,222 images of a small number of people with eight different mask usage annotations. The Medical Mask Dataset (MMD)~\cite{mmd} and AIZOOTech dataset~\cite{AIZOOTech} are publicly available masked face detection datasets and annotation efforts published by private  companies. 

\section{Open Issues \& Future Challenges}\label{Sec5}
Significant progress has been made over the last couple of years to address the main challenges in CVHA techniques for the COVID-19 era, but several open issues still remain and need to be addressed in the future. Below we provide a discussion of the most important topics in the opinion of the authors.

\subsection{Self-adaptation of CVHA techniques}

A desired ability of CVHA techniques is detecting new conditions and self-adaptation to these new conditions. For example, before the COVID-19 era it was not very common to wear masks. Therefore, most of the pre-COVID-19 CVHA techniques were trained with datasets that do not at all contain or contain very few samples of people with masks. On account of this, the performance of many existing techniques deteriorated severely, once people started wearing face masks. As the changes in human appearance can occur over time due to several factors, e.g., due to fashion trends and health concerns, it is necessary and important to have CVHA techniques that adapt themselves to the current conditions. One way of performing this is to benefit from online continual learning approaches~\cite{li2017learning, lopez2017gradient, shin2017continual, aljundi2019task, parisi2019continual, van2019three} that have been utilized in computer vision research. Detecting unseen cases is critical, as sometimes changes in appearance might not be as sudden. 
The adaption of models to new concepts and conditions by learning with few data~\cite{vinyals2016matching, gidaris2018dynamic, yin2020dreaming, zhang2021few} and using self-supervised features~\cite{ahmad2022few} are other essential points that need to be considered in future works.

\subsection{Generalization and robustness}

The current generation of CVHA techniques is already exhibiting remarkable performance across diverse data characteristics. However, in unconstrained scenarios, large appearance variability may still adversely affect their performance~\cite{Wang21,alzu2021masked}. This, for example, includes low-resolution masked inputs for tasks such as face landmarking, face detection and face recognition. Significant pose variations and additional occlusions, e.g., due to glasses, hats and scarfs, also still have an adverse effect on performance, especially with masked facial images. In CVHA solutions involving crowds, novel ideas and powerful techniques are needed that can differentiate between masked and non-masked people in various environments and across a range of viewing angles~\cite{nguyen2021effectiveness}. Thus, there is an imminent need to further improve the generalization capabilities and most of all robustness of CVHA techniques aimed at combating COVID-19.     

\subsection{Availability of large-scale benchmarks}

As discussed in Section~\ref{Sec3}, a considerable amount of datasets, especially for the masked faces, appeared in response to the needs induced by the COVID-19 pandemic. However, many of these datasets are small, not well curated and come without a well-defined experimental protocol and/or performance indicators. In CVHA problems, such as facial landmarking and related tasks, for example, ground truth information is usually also not readily available. As a result, research often combine datasets for their experiments and define in-house protocols for experimentation. It is, therefore, difficult to objectively evaluate progress and assess the merits and deficits of the CVHA techniques being proposed in the literature. Well-designed large-scale benchmarks with clear objectives and properly designed experimental protocols are critically needed to help advance the field further and provide a solid basis for research going forward.      

\subsection{Bias and fairness}

Data-driven techniques that learn from labeled examples are today the most widely utilized solutions for various computer vision tasks. When such techniques are applied in automated decision-making systems that impact people's lives, fairness and bias become critically important. As automated decisions need to be fair and equally accurate for all, regardless of race, gender, age and other demographic factors, it is paramount that CVHA techniques ensure unbiased performance for subjects with diverse demographic attributes~\cite{drozdowski2020demographic}. 
The negative consequences of biased systems have, for example, made headlines for face recognition, prompting many of the largest software corporations, such as Microsoft, Amazon, and IBM, to reconsider their face recognition programs and policies~\cite{Heilweil2020,meden2021privacy}. While several studies explored bias with standard CVHA techniques~\cite{puc2021analysis,robinson2020face,cavazos2020accuracy,albiero2021gendered,babic2020does}, this issue has seen far less attention with masked face images and the data characteristics induced by COVID-19~\cite{yu2021boosting}. Therefore, studies are needed that help to better understand the  behavior of CVHA techniques in terms of bias and fairness with masked face images, as well as targeted mitigation techniques that contribute towards fairer decisions for different CVHA tasks. Furthermore, as many of the existing datasets gathered for COVID-19 related CVHA techniques are not balanced across demographic groups, additional efforts are also required on the data collection and curation side to facilitate research into these topics. 

\subsection{Ethics and privacy}

The ability of processing face images behind masks and identifying people raises certain questions about the surveillance capabilities enabled by such technologies. As in all ethics issues, a sound analysis should balance the potential benefits against the potential risks, and arrive at guidelines and recommendations that will mitigate the risks, while maximizing the benefits. The main risks of facial surveillance involve loss of privacy, especially in cases where privacy matters. Government surveillance, in particular during events that criticize the said government, is a major case in point, and there is widespread worry that the deployment of facial surveillance can jeopardize people's rights of expression, can lead to prosecutions and harm. This is also linked to legitimate uses of facial surveillance and analysis technology, such as for public transport payments or health related screening purposes, which can then be extended into citizen surveillance, i.e. the ``slippery slope'' argument. Religious freedoms, freedom of opinion and expression, freedom of assembly and association are all fundamental human rights, and need to be protected. During the \textit{Umbrella Movement} in Hong Kong, the protesters have used masks and other props to cover their faces to prevent the police from identifying them and singling out protesters for arrests~\footnote{\url{https://www.nytimes.com/2019/07/26/technology/hong-kong-protests-facial-recognition-surveillance.html}}. While there may be legal barriers for governments to target protesters, they can be sidestepped quickly. A Chinese based company, Hanwang, announced in 2020 that its facial recognition software was identifying people with masks with 95\% accuracy, as opposed to 99.5\% for people without masks~\footnote{Martin Pollard, ``Even mask-wearers can be ID'd, China facial recognition firm says,'' Reuters, 9 March 2020, retrieved from \url{https://reut.rs/2TAwMux }}. When asked the possibility of this software being used to identify protesters in Hong Kong, the company spokesperson said that this use case is known, but the market is too small. The company reports having about 200 clients in Beijing using the technology, including the police.

The second potential risk for enhanced facial surveillance capabilities (even with masked faces) is associated with data use by private companies. Since identification and personalization content for potential customers is key for new marketing approaches, identity can be monetized easily. Privacy breaches in this sector can have drastic consequences\footnote{``Before Clearview Became a Police Tool, It Was a Secret Plaything of the Rich,'' The New York Times, March 2020, \url{https://www.nytimes.com/2020/03/05/technology/clearview-investors.html}}.
While personal data, including biometric data, such as facial imagery, are regulated in certain parts of the world, e.g., see GDPR in Europe\footnote{Accessible from: \url{https://gdpr-info.eu/}}, the Japanese Act on the Protection of Personal Information in Japan~\cite{nec}, or the California Consumer Privacy Act (CCPA)~\cite{10.2307/j.ctvjghvnn} and the Biometric Information Privacy Act (BIPA)~\cite{bipa} in the US, technological safeguards are also critically needed to address ethics and privacy concerns. Along these lines, biometric privacy-enhancing technologies designed specifically for masked faces and capable of hiding part of the information contained in the data may become more important going forward~\cite{meden2021privacy}.

\section{Conclusion}

It is generally expected that COVID-19 will not simply disappear, and will remain an issue for years to come. Consequently, novel computer vision techniques adapted to societal developments and behavioral changes of people induced by prevention measures and health-related governmental policies will increasingly be needed. A significant amount of work has already been done to help prevent and control the spread of the disease and facilitate normal operation of identity management schemes and other relevant infrastructure using vision-based methods. As discussed in the survey paper, a large part of this work focused on computer vision techniques for human analysis (CVHA), which analyze visual data related to faces and people during the COVID-19 era, e.g., in the presence of occlusions with face masks.

In this survey paper, we presented a comprehensive review of existing CVHA solutions for the COVID-19 era. Specifically, we discussed the main challenges introduced to CVHA problems by the pandemic, presented a high-level taxonomy of existing methods, elaborated on relevant datasets and described, what we feel, are the most important open issues that need to be addressed in the future. The consolidated information presented in the survey is expected to help researchers working on similar problems to quickly get an overview of the work already done and the main challenges that require further research.

\section*{Acknowledgements}

This research was supported in parts by the ARRS Research Programme P2–0250 (B) ``Metrology and Biometric Systems'' and the additional funding provided for COVID-19 related research as well as the bilateral ARRS-TUBITAK funded project: Low Resolution Face Recognition (FaceLQ), with TUBITAK project number 120N011. 
This research work has been also partially funded by the German Federal Ministry of Education and Research and the Hessen State Ministry for Higher Education, Research and the Arts within their joint support of the National Research Center for Applied Cybersecurity ATHENE, and EU COST project GoodBrother (19121).
The project on which this report is based was also partially funded by the Federal Ministry of Education and Research~(BMBF) of Germany under the number~01IS18040A. 

\bibliographystyle{elsarticle-num}
\bibliography{mybibfile}

\end{document}